# Interpretable Similarity of Synthetic Image Utility

Panagiota Gatoula, George Dimas, and Dimitris K. Iakovidis, *Senior Member, IEEE*

***Abstract*— Synthetic medical image data can unlock the potential of deep learning (DL)-based clinical decision support (CDS) systems through the creation of large scale, privacy-preserving, training sets. Despite the significant progress in this field, there is still a largely unanswered research question: "How can we quantitatively assess the similarity of a synthetically generated set of images with a set of real images in a given application domain?". Today, answers to this question are mainly provided via user evaluation studies, inception-based measures, and the classification performance achieved on synthetic images. This paper proposes a novel measure to assess the similarity between synthetically generated and real sets of images, in terms of their utility for the development of DL-based CDS systems. Inspired by generalized neural additive models, and unlike inception-based measures, the proposed measure is interpretable (Interpretable Utility Similarity, IUS), explaining why a synthetic dataset could be more useful than another one in the context of a CDS system based on clinically relevant image features. The experimental results on publicly available benchmark datasets from various color medical imaging modalities including endoscopic, dermoscopic and fundus imaging, indicate that selecting synthetic images of high utility similarity using IUS can result in relative improvements of up to 54.6% in terms of classification performance. The generality of IUS for synthetic data assessment is demonstrated also for grayscale X-ray and ultrasound imaging modalities. IUS implementation is available at *https://github.com/innoisys/ius*.**

## I. INTRODUCTION

THE adoption of Deep Learning (DL) models for Clinical Decision Support (CDS) in biomedicine is still limited. Main causes include their "black box" nature, which affects users' trust, data privacy and annotation cost that bounds their generalization capacity. Moreover, the deployment of DL models for CDS depends on the availability of training data. Obtaining suitable datasets is often complicated due to restrictive sharing policies, the absence of comprehensive annotations, class imbalances, and inconsistent dataset sizes[1]. Explainable/interpretable approaches to DL and synthetic medical image generation methods are contemporary approaches tackling such issues. Regarding medical image synthesis, deep generative models based on Generative Adversarial Networks (GANs) [2], Variational Autoencoders (VAEs) [3], and more recently Diffusion Models [4], provide state-of-the-art performance [5–8]; Generative approaches in medical imaging demonstrate a wide range of applications [1], including: (a) data augmentation, *i.e.,* leveraging synthetic data to expand real training datasets; (b) image reconstruction and enhancement, *i.e.,* generating synthetic data to improve image resolution or quality; and (c) medical image translation, *i.e.,* synthesizing images to convert real images of a source modality to a target modality (*e.g.*, CT-to-MRI conversion). Additionally, recent studies have reviewed from a clinical perspective, synthetic medical images to assess their potential in supporting clinical tasks [9], [10]. However, the evaluation of synthetic medical images, and particularly the assessment of their utility in DL applications, remains an area that has not been thoroughly investigated [1].

The extent to which the synthetic data can substitute the original data without materially altering the analytic outcomes derived from it, is referred to as synthetic data utility. Utility is not always correlated with fidelity, which refers to the degree the generated images resemble the real ones, in the sense that they maintain the essential distributional characteristics of the original dataset, *e.g.*, in the context of CDS, a set of higher fidelity synthetic images may have lower utility if they are not as diverse as the respective real dataset. Diversity is a complementary quality that measures whether the generated images cover the full variability of the real dataset [11]. A high-fidelity dataset exhibiting a similar level of diversity with its real counterpart is expected to have a similar utility Today, the utility assessment of synthetic images for CDS is mainly addressed via task-specific performance measures, such as classification accuracy. For example, assessing if real images can be classified with the same accuracy whether the classifier is trained on respective real or synthetic data [1]. The quality assessment of synthetic images in terms of fidelity and/or diversity, usually relies on user evaluation studies [9], [12], and inception-based measures originally developed for non-medical images, such as the Inception Score (IS) [13], and the Fréchet Inception Distance (FID) [14]. However, the applicability of such measures in medical imaging is regarded controversial [15], mainly because they are based on DL models (*e.g.*, InceptionV3 [16]) pretrained on non-medical datasets *i.e.,*

This work is part of the European project SEARCH (https://ihi-search.eu/), which is supported by the Innovative Health Initiative Joint Undertaking (IHI JU) under grant agreement No. 101172997. The JU receives support from the European Union's Horizon Europe research and innovation programme and COCIR, EFPIA, Europa Bio, MedTech Europe, Vaccines Europe, Medical Values GmbH, Corsano Health BV, Syntheticus AG, Maggioli SpA, Motilent Ltd, Ubitech Ltd, Hemex Benelux, Hellenic Healthcare Group, German Oncology Center, Byte Solutions Unlimited, AdaptIT GmbH. Views and opinions expressed are, however, those of the author(s) only and do not necessarily reflect those of the aforementioned parties. Neither of the aforementioned parties can be held responsible for them. *(Corresponding author: Dimitris K. Iakovidis).*

P. Gatoula, G. Dimas, and Dimitris K. Iakovidis are with the University of Thessaly, Department of Computer Science and Biomedical Informatics, Lamia 35131 Greece (e-mail: {pgatoula, gdimas, diakovidis}@uth.gr).

This is a preprint of an article published in IEEE Transactions on Medical Imaging, vol. 45, no. 7, pp. 3529–3544, July 2026, doi: 10.1109/TMI.2026.3679527. The final published version is available at https://doi.org/10.1109/TMI.2026.3679527.

TABLE I
SUMMARY OF CURRENT MEASURES FOR SYNTHETIC MEDICAL IMAGE ASSESSMENT

| Measures | Characteristics | | | |
|---|---|---|---|---|
| | Image-level evaluation | Distribution-level evaluation | DL-utility assessment | Interpretability |
| *Image similarity measures, e.g., MAE, SSIM, PSNR* | ✓ | - | - | - |
| *CNN-based feature similarity measures, e.g., IS, FID, KID, LPIPS* | - | ✓ | - | - |
| *Task performance measures, e.g., classification AUC/Accuracy* | ✓ | - | ✓ | - |
| ***IUS (proposed)*** | ✓ | ✓ | ✓ | ✓ |

ImageNet [17], and they usually fail to capture clinically relevant features of anatomical structures and pathological details due to the large domain shift [18]. Such measures also evaluate distributions of synthetic images based on features extracted from deeper layers of DL models; thus, failing to assess the quality of individual synthetic images and neglecting image-specific details essential for precise description of individual images. Moreover, existing inception-based measures lack interpretability, a critical aspect for addressing the clinical relevance of medical images. They rely on black box DL models lacking transparency regarding their evaluation criteria.

Recent studies indicate the potential for interpreting clinical features within medical images using Generalized Neural Additive Models (GAMs) [19], [20]; hence, coping with the lack of transparency that characterizes regular DL models. Motivated by these findings, this paper introduces the concept of *utility similarity*, which to the best of our knowledge, is used for the first time. *Utility similarity* is defined as a measure for assessing how similar a dataset of synthetic images is, with a dataset of real images in terms of utility (usefulness) in DL tasks [1] and, it is applied in the context of medical image classification. Furthermore, the proposed utility measure is interpretable, offering a way to explain why a synthetic dataset could be more useful than another one based on image features that are usually clinically-relevant, *i.e.*, visually assessed by clinicians for clinical decision making, such as color and texture [10]. The proposed *Interpretable Utility Similarity* (IUS) achieves that by considering the interpretable responses of a generalized additive Convolutional Neural Network (CNN) following the E Pluribus Unum CNN (EPU-CNN) framework, which is guided by selected, perceptual feature representations. Advantages of IUS as a tool in medical imaging, include: (a) it is unaffected by domain shift caused from training on non-medical datasets, such as ImageNet, since by definition it assesses the similarity of images that belong to the same domain. (b) Contrary to contemporary measures, that only perform distribution-level assessments, IUS enables the evaluation of singletons, *i.e.*, individual images; (c) it yields estimations aligned with the classification performance of DL models assessing the utility of synthetic data; (d) it constitutes a general approach mitigating the need for such exploratory classification experiments and arbitrary, domain-specific, classification performance comparisons by providing a standard scale to assess the utility of the synthetic datasets.

The rest of this paper is structured as follows: Section II presents a literature review on current approaches. Section III describes the proposed IUS measure. Section IV showcases experimental results, and the last section discusses the conclusions derived from this study and directions for future research.

## II. RELATED WORK

Today the most reliable approach to assessing the similarity of synthetic medical images with their real counterparts is performing time-consuming and costly user studies, requiring engagement from multiple experts [1]. This approach is based on perceived, qualitative criteria that are usually subjective. Despite current advances in synthetic data generation, objective similarity assessment remains a challenge due to the lack of standards and trustworthy quantitative measures [5], [21].

Table I summarizes the existing measures for synthetic image assessment, highlighting their main characteristics regarding their applicability level (*i.e.,* applicable per image or distribution-wide), interpretability, and relevance to utility with respect to DL. Image similarity measures, such as Mean Absolute Error (MAE), Peak Signal-to-Noise Ratio (PSNR), Structural Similarity Index (SSIM) *etc.*, are applicable to supervised synthesis tasks, (*e.g.,* super-resolution, image-to-image translation) [1]. These are image-level measures, typically comparing pixel-wise or regional differences between synthetic and real images, yet they lack interpretability and utility assessment. CNN-based feature similarity measures, such as FID, IS, Kernel Inception Distance (KID) and Learned Perceptual Image Patch Similarity (LPIPS) [13], [14], [22], [23] compare higher-level image features, *i.e.*, semantically more meaningful representations, rather than focusing on low-level pixel comparisons [1]. They rely on CNNs pretrained on large-scale natural image datasets (*e.g.,* ImageNet) to extract high-level feature representations and evaluate the distributional similarity between real and synthetic image sets in a learned embedding space. Specifically, FID and KID assess the distributional similarity between real and synthetic images based on feature statistics. IS evaluates the quality and diversity of synthetic images based on classification confidence. LPIPS evaluates visual similarity between real and synthetic images leveraging deep feature activations. However, such measures lack interpretability. While alternative variants of these measures, *e.g.,* modified IS, have been proposed to address performance limitations [15], they are tailored for non-medical images. Therefore, these adaptations still evaluate synthetic images on a distribution-level rather than individually, limiting their applicability in the medical context.

Despite their limitations, Inception-based measures, *i.e.,* feature similarity measures using InceptionV3 CNN, such as FID, IS and KID, remain widely used in synthetic medical image assessment. Nevertheless, their reliability is still questioned because they are pretrained on non-medical datasets [15], [24]. To cope with that, previous studies have investigated the impact of medical pretraining strategies for such inception-based measures [25]. However, the results obtained did not lead to reliable estimations or improvements upon conventional inception-based scores.

Performance measures assess synthetic image sets by monitoring their impact in the deployment of CDS systems trained for specific DL tasks [1]. To this end, classification models are trained using synthetic data and then they are evaluated on real data, employing measures dependent on the task performed, such as accuracy, or the Area under the receiver operating Characteristic (AUC). While these measures infer the overall utility of synthetic image sets incorporated in DL-predictive tasks, they do not quantify the similarity of individual images nor provide interpretability regarding the source of the observed utility.

The proposed IUS measure evaluates the similarity of synthetic images with respect to their usefulness for effective training of DL-based CDS systems. Unlike performance measures that assess the overall utility of synthetic image sets through time-consuming experiments like image classification, IUS can provide scores for individual images, indicating their utility similarity to a real image set, incorporated in DL-predictive tasks. Furthermore, IUS relies on transparent evaluation criteria, such as color and texture, that are perceptually relevant to clinical image interpretation [10], [19], contrary to inception-based measures that rely on black-box approaches; therefore, it can be considered as more trustworthy in that sense. Additionally, IUS provides interpretable results regarding its evaluation criteria.

## III. METHODOLOGY

The methodology for the estimation of the IUS measure leverages a perceptually interpretable EPU-CNN model. Initially, this model is trained so that it learns to classify real images from a medical domain of interest. During training, the EPU-CNN model learns to interpret the classification outcomes in terms of perceptual features characterizing the image content, such as texture and color. IUS can be used to assess if a set synthetic of synthetic images is as useful as a real set of images to train a classifier that implements an image-based CDS system. An outline of the utility similarity assessment of synthetic images using IUS is illustrated in Fig. 1. Its estimation involves three steps: a) Initially, the trained EPU-CNN model is applied on a real reference image set, which infers not only the class membership of the images, but also interpretations in the form of a vector with components explaining which perceptual features contributed to inferring that membership. By averaging these interpretations, a baseline feature contribution profile, $\boldsymbol{C_b}$ is acquired. This profile describes the positive (depicted in Fig. 1 with green color bars) or negative (depicted in Fig. 1 with red color bars) contribution of the perceptual features of the image to the EPU-CNN classification outcome. The construction of the baseline feature contribution profile can be estimated offline. b) Then, the trained EPU-CNN model can be fed with synthetic images. The interpretations obtained from the inference process form a feature contribution profile $\widetilde{\boldsymbol{C}}$, tailored for each synthetic sample. c) To assess the utility similarity of each synthetic image to a real image set based on interpretable evaluation criteria, the similarity between $\widetilde{\boldsymbol{C}}$ and $\boldsymbol{C_b}$ is calculated. The resulting IUS score can be mapped to a qualitative utility level, *e.g.*, Very High, indicating the usefulness of individual synthetic images in DL tasks (Fig. 1). These steps are explained in detail in the following paragraphs.

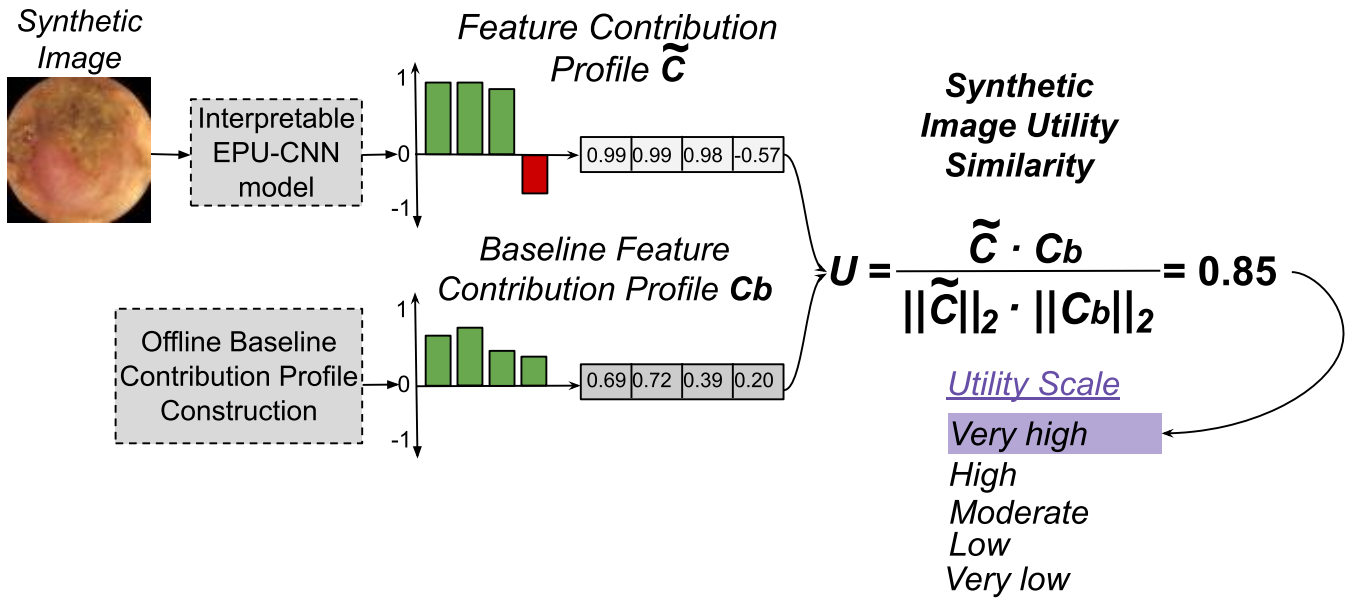

Fig. 1. Synthetic image assessment using IUS.

### A. EPU Interpretable Image Classification

The EPU-CNN model constitutes the basis for the calculation of the IUS measure. It has the following formulation:

$$\mathbb{E}\left[Y|\boldsymbol{I};\theta_{\{N\}}\right] = \frac{1}{1+e^{-\left(\boldsymbol{b}+\sum_{i=1}^{N} f_i(I_i;\vartheta_i)\right)}} \quad (1)$$

where $\boldsymbol{b}$ is a trainable bias parameter, $\mathbb{E}\left[Y|\boldsymbol{I};\theta_{\{N\}}\right]$ is the expected value of ground truth $Y$ given a set of perceptual feature maps (PFMs) $\boldsymbol{I} = (I_1, I_2, \ldots, I_N)$. The PFMs derive from a decomposition process of an image $I$ [19]. Each function $f_i(\cdot\,;\vartheta_i), i = 1,2,\ldots,N,$ denotes a sub-network parametrized by $\vartheta_i$ that learns a mapping $I_i^{H\times W\times C} \rightarrow I^L$, with $H$, $W$ and $C$ denoting the height, width and the number of channels of a PFM $I_i$. $L$ indicates the dimensionality of the output space which depends on the number of classes involved in the CDS application, *e.g.*, for a binary decision-making problem $L=1$. It is worth noting that each $f_i$ is used to process only a specific PFM. The output layer of each sub-network utilizes the hyperbolic tangent (tanh) activation [19], [20], yielding an output within the range of [–1, 1].The final prediction $\mathbb{E}\left[Y|\boldsymbol{I};\theta_{\{N\}}\right]$ balances all outputs allowing complementary or conflicting cues across PFMs to influence the final prediction. The EPU-CNN formulation ensures that perceptually relevant feature interactions are reflected in the final prediction [19]. A vector composed of the output of all functions $f_i, i = 1,2,\ldots,N,$ forms the interpretation provided by the EPU-CNN model, encoding the contribution of each PFM, $I_i$, to the classification outcome. This interpretation vector forms the feature contribution profile of an image $I$ to the class membership. Each $f_i$ response indicates how strongly each perceptual feature *i.e.,* PFM, aligns (positive contribution) or opposes (negative contribution) to the class membership. Thus, image classification is interpreted in terms of perceptual features via the feature contribution profile.

This paper focuses on color medical images as a proof-of-concept approach adopting the original EPU-CNN approach documented in [19]. In that approach four PFMs are considered to represent orthogonal opponent image features encoding color and texture: (a) green-red (G-R) and (b) blue-yellow (B-Y) color, and (c) light-dark (L-D), and (d) coarse-fine (C-F) texture. G-R and B-Y PFMs are the *a* and *b* components of the CIE-*Lab* color-space, and the L-D and C-F PFMs are estimated through a three-level 2D discrete wavelet transform (DWT) on the *L* component of that space. These PFMs serve two critical roles: (a) they align with established human opponent-process mechanisms (ensuring that each channel encodes independent information), and (b) they reduce the redundancy that is inherent in the RGB color space, improving robustness and CNN efficiency [19].

In practice, the color PFMs, capture hue antagonisms. The multi-resolution luminance PFMs, and C-F, reflect the sensitivity of the human eye to different spatial frequencies. In detail, the L-D PFM represents the low-frequencies of an input image that usually encode coarse shapes, *e.g.*, distinguishing broad organ contours from surrounding tissue. C-F represents the spectrum of high-frequencies encoding texture and fine details. Similarly to [19], in case of color medical images all four PFMs are used, whereas for grayscale medical images, *e.g.*, MRI scans, only the L-D and C-F PFMs can be used [20].In this paper we focus on color medical images since they usually include subtle tissue contrasts, making synthetic data more challenging to generate and valuable for clinical decision support [26].

The training of the EPU-CNN model is performed in an end-to-end manner by minimizing an entropy-based loss $\mathcal{L}$:

$$\theta_{\{N\}} = argmin_{\theta}\ \mathcal{L}(Y, \mathbb{E}\left[Y|\boldsymbol{I};\theta_{\{N\}}\right]) \tag{2}$$

where, $\mathcal{L}(Y, \mathbb{E}\left[Y|\boldsymbol{I};\theta_{\{N\}}\right])$ is formulated as:

$$-\frac{1}{L}\sum_{\ell=1}^{L} y_{\ell}\log\left(\mathbb{E}\left[y_{\ell}|\boldsymbol{I};\theta_{\{N\}}\right]\right) \tag{3}$$

in the case of *L=1, i.e.,* binary classification, Eq. (3) is modified:

$$-y\log\left(\mathbb{E}\left[y|\boldsymbol{I};\theta_{\{N\}}\right]\right) - (1-y)\log\left(1-\mathbb{E}\left[y|\boldsymbol{I};\theta_{\{N\}}\right]\right) \tag{4}$$

It is worth noting that, while the EPU-CNN framework combines sub-network outputs additively, it effectively accommodates complex feature interactions in two ways [19]. First, each CNN sub-network, is inherently capable of learning hierarchical, non-linear spatial patterns from its respective PFM. Subsequently, the summation process of all sub-network responses, weights evidence across the sub-networks. This allows complementary or conflicting contributions deriving from different PFMs to influence the final prediction. Second, due to the joint training of the sub-networks of the EPU-CNN model, relationships among them are implicitly re-parameterized during back-propagation, so that they have complementary contributions. Together these mechanisms ensure that high-level feature interactions are reflected in the final prediction of the EPU-CNN model [19]. EPU-CNN constitutes an interpretable classification framework capable of consistently quantifying the contribution of perceptual image features to the final classification outcome regardless of the underlying sub-network architecture. Furthermore, this framework exhibits training stability using various architectures [19]. These aspects render it robust for the calculation of the IUS measure.

## B. Utility Similarity of Synthetic Medical Images

As a first step the EPU-CNN model is trained on a real image dataset. To this end, each real image, *I*, is decomposed to four PFMs described in Section III.A. Once the EPU-CNN model is trained, the output of each sub-network $f_i(I_i;\vartheta_i)$ can provide interpretations for the classification outcome of $I_i$ encoding the contribution of the PFMs, representing perceptual features to the class membership. These interpretations are represented by $\boldsymbol{C}$, denoting the feature contribution profile of an image *I*:

$$\boldsymbol{C} = [f_1(I_1;\vartheta_1), \dots, f_N(I_N;\vartheta_N)] \tag{5}$$

Thus, by propagating an individual synthetic image, $\tilde{I}$, or a batch of synthetic images to the EPU-CNN model, the perceptual feature profile of an image(s) can be quantified to intuitively interpret the class membership w.r.t. the perceptual features. The feature contribution profile of an individual synthetic image is used for the assessment of its utility similarity to a real image set. Each feature contribution profile consists of the PFM contributions estimated by the EPU-CNN, forming a vector $\widetilde{\boldsymbol{C}} = \left[f_1\left(\widetilde{I_1};\theta_1\right), \dots, f_N\left(\widetilde{I_N};\theta_N\right)\right]$. IUS expects that clinically accurate images should exhibit feature-contribution profiles resembling a baseline profile, $\boldsymbol{C_b}$, derived from real images. Given a set of *J* real images, *I*, either labeled or unlabeled, that were not included in the training set, *e.g.*, a test set, $\boldsymbol{C_b}$ is computed offline by averaging all vectors $\boldsymbol{C_i}$ (Eq. (5)) constituting the interpretations of EPU-CNN for each real image *I*:

$$\boldsymbol{C_b} = \frac{1}{J}\sum_{i=1}^{J}\boldsymbol{C}_i \tag{6}$$

A similarity function is employed to calculate the similarity between a feature contribution profile $\widetilde{\boldsymbol{C}}$ that corresponds to a synthetic image and the baseline feature contribution profile, $\boldsymbol{C_b}$. To this end, the cosine similarity function was selected because it captures the directional, *i.e.*, positive or negative, relationships between vectors. Furthermore, it has been effectively applied in various clinical evaluation contexts to compare feature vectors and embeddings, capturing essential directional relationships between them [27], [28]. The values of cosine similarity constrain the utility similarity scores (IUS scores) within a standard scale ranging in the interval of [-1, 1]. Scores closer to 1 indicate a higher similarity between $\widetilde{\boldsymbol{C}}$ and $\boldsymbol{C_b}$, whereas values closer to -1 indicate a higher dissimilarity between them. Scores near to 0 indicate orthogonality between them, *i.e.*, a lack of correlation between them. Hence, the IUS score for a synthetic image can be assessed according to the following equation:

$$\mathcal{U} = \frac{\widetilde{\boldsymbol{C}}\cdot\boldsymbol{C_b}}{\left\|\widetilde{\boldsymbol{C}}\right\|_2\cdot\|\boldsymbol{C_b}\|_2} \tag{7}$$

where $\|\cdot\|_2$ denotes the Euclidean norm. By estimating the average feature contribution profile of a synthetic image set, $\widetilde{\boldsymbol{C}}_{avg}$, Eq. (7) can be used to directly estimate the utility of a synthetic medical image dataset by replacing $\widetilde{\boldsymbol{C}}$ of a singleton with $\widetilde{\boldsymbol{C}}_{avg}$.

## IV. Experiments and Results

### A. Experimental Setup

A comprehensive evaluation of the IUS measure was conducted on six publicly available medical benchmarks, typically considered in image synthesis studies, including endoscopic, dermoscopic and retinal image datasets [29–34] (Table II). ISIC includes normal and abnormal instances that were considered pairwise, *i.e.*, melanoma (Me) vs. nevi (Ne), basal cell carcinoma (Ca) vs. Ne, and Me vs. Ca, in line with [19]. Moreover, X-ray and ultrasound datasets from the MedMNIST data collection [35] were used to further assess the generalization of IUS beyond color imaging (Section IV.G). To extend the assessment of IUS in the context of synthetic image selection for endoscopy video classification, PolypGen [36] was selected as one of the most comprehensive ones. Overall, the selection of datasets was based on their popularity as benchmark datasets in medical imaging, and their compliance to the 'Findable, Accessible, Interoperable and Reuse' (FAIR) principle [37] (Section IV.H).

A set of EPU-CNN models was trained with these datasets to assess the feature contribution profiles that are used by IUS to estimate the utility similarity of synthetic images to real image sets. The $Base_I$ architecture, that has been proposed in the context of EPU-CNN framework [19], was used as a base sub-network as it provides balance between performance and complexity. Each EPU-CNN was trained using the Stochastic Gradient Descent algorithm with an initial learning rate of $1\times10^{-3}$, momentum 0.9, and with a batch size of 64. The early stopping mechanism was adopted to prevent overfitting. The proportions of training, validation and testing data splits were 70%, 20% and 10%, respectively. The training of the EPU-CNN was performed on a workstation equipped with an AMD Ryzen 7 5800X 8-Core Processor (boost clock up to 4.85GHz) and a single NVIDIA RTX 3080 GPU with 12GB VRAM. The EPU-CNN framework contains $15.34\times10^6$ trainable parameters. To quantify its computational complexity, we measured the number of floating-point operations (FLOPs) required to process a single image, yielding $24.28\times10^9$ FLOPs. During inference, the computational overhead includes the propagation of a synthetic image $\tilde{I}$ through the EPU-CNN, the estimation of its feature contribution profile $\widetilde{\boldsymbol{C}}$, and the computation of IUS score. In terms of execution performance, the computation of IUS for 10,000 synthetic images, on the utilized hardware, requires an average of $11.87\times10^{12}$ FLOPs per second (FLOPS). This corresponds to a throughput of approximately 489.2 images per second. As previously described, the baseline feature contribution profile $\boldsymbol{C_b}$ is estimated offline and therefore does not affect the inference time.

To facilitate the evaluation of IUS measure, several synthetic images were generated, based on the datasets of Table II, using two state-of-the-art synthetic image generation methods: (a) the StyleGANv2 model (*Generator-I*) [38] and (b) a variational autoencoder inspired by the architectural design of Visual Transformers - ViTs (*Generator-II*) [10]. These models were selected due to their stability even on smaller medical datasets [5], [25], enabling time-efficient generation of large, consistent pools of synthetic images, as a solid basis for the systematic evaluation of IUS. Although diffusion models are also a contemporary option, their longer sampling times and higher computational needs make them an unattractive option for conducting a large-scale evaluation process of multiple training and inference instances [39]. It is also important to underline that this work aims at assessing the capacity of IUS in measuring the utility similarity of synthetic medical images to real images in the context of DL applications; hence, any comparison regarding the output quality of the generative models, falls beyond the scope of this work.

### B. Utility Analysis

To investigate the effectiveness of IUS for selecting synthetic images of high utility similarity to real image sets, and its relation with relevant measures widely adopted in the state of the art, we performed comparative experiments using the generated images for various IUS interval scores, hereinafter IUS scores, with those measures. For this analysis the InceptionV3 CNN was used. It was trained using the Adam Optimizer with a learning rate of $1\times10^{-4}$. The batch size was set to 128 and the early stopping strategy was applied to prevent

TABLE II
Datasets Used for the Experiments

| Dataset | Modality | Samples | Categories |
|---|---|---|---|
| KID [29] | Capsule Endoscopy | 2,371 | Inflammatory, vascular polypoid lesions, and normal mucosa from esophagus, stomach, small-bowel, colon. |
| Kvasir-Capsule [30] | Capsule Endoscopy | 37,370 | Angiectasias, erosions, erythemas, lymphangiectasias, polyps, ulcers; normal tissue. |
| Kvasir-v2 [31] | Flexible Endoscopy | 6,000 | Esophagitis, polyps, ulcerative colitis findings; normal mucosa from z-line, pylorus and cecum anatomical landmarks. |
| ISIC 2019 [34] | Dermoscopy | 9,000 | Skin lesions including melanoma (Me), melanocytic nevus (Ne), basal cell carcinoma (Ca), squamous cell carcinoma (Ca). |
| APTOS 2019 [33] | Retinal Fundus | 1,600 | Mild, moderate and severe non-proliferative diabetic retinopathy (DR), proliferative DR, and non-DR. |
| DeepDRiD [32] | Retinal Fundus | 3,662 | Mild, moderate and severe non-proliferative DR, proliferative DR, and non-DR. |
| BreastMNIST [35] | Ultrasound | 780 | Malignant and normal tissue (including benign lesions). |
| PneumoniaMNIST [35] | X-ray | 5,856 | Pneumonia and normal lung parenchyma. |
| PolypGen [36] | Video Colonoscopy | 8,037 | Polypoid lesions and normal mucosa. |

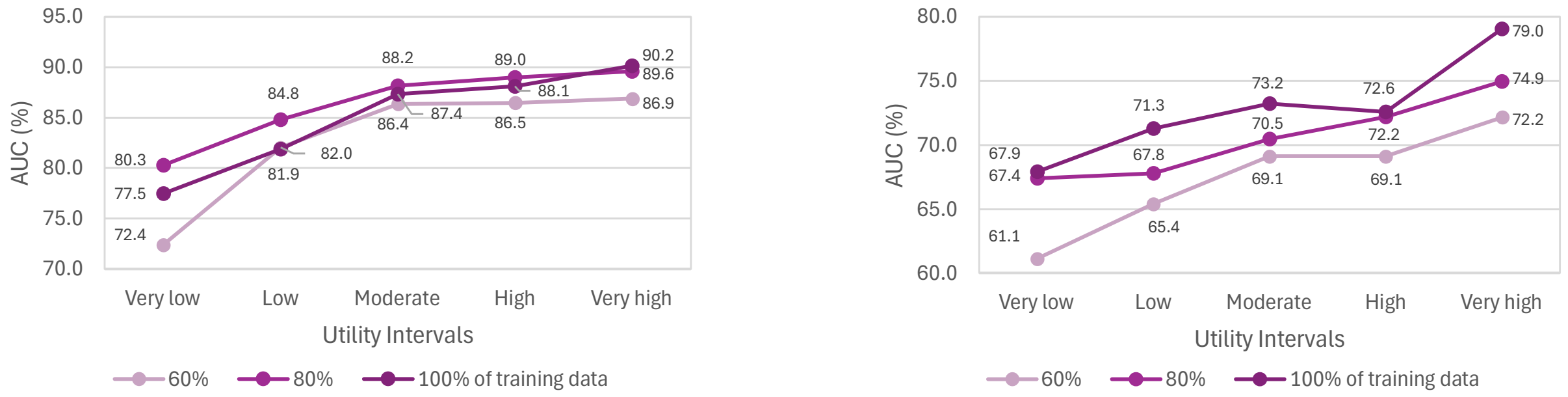

(a) (b)

Fig. 2. InceptionV3 CNN classification performance on models trained solely with synthetic images of varying IUS score, produced by (a) Generator-I and (b) Generator-II synthesis methods and evaluated on real images.

TABLE III
INCEPTION-BASED MEASURES FOR DIFFERENT IUS LEVELS

| IUS | *Generator-I* | | *Generator-II* | |
|---|---|---|---|---|
| | ***FID*** ↓ | ***IS*** ↑ | ***FID*** ↓ | ***IS*** ↑ |
| *Very low (VL)* | 58.6 | 1.9 | 70.5 | 2.2 |
| *Low (L)* | 54.2 | 2.0 | 65.4 | **2.3** |
| *Moderate (M)* | 51.7 | 2.1 | 53.0 | 2.2 |
| *High (H)* | 50.2 | **2.1** | 45.2 | 2.2 |
| *Very high (VH)* | **46.9** | 2.0 | **39.0** | 2.2 |

↑ and ↓ indicate that higher and lower is better, respectively.

overfitting. The training validation and testing sets were split into 70%, 20%, and 10%, respectively. Stratified five-fold cross validation was used to mitigate data selection bias.

The range of IUS scores was divided into five intervals. Specifically, four threshold values (T = 0.2, 0.4, 0.6, 0.8) resulting in five non-overlapping intervals expressing the degree of utility similarity between synthetic and real image sets, *i.e.*, [-1.0, 0.2), [0.2, 0.4), [0.4, 0.6), [0.6, 0.8), and [0.8, 1.0], were considered. These intervals correspond to a five-level scale. Samples with negative IUS scores were included in the lowest interval due to low occurrence, *i.e.*, the interval [-1, 0.2) represents synthetic images of Very Low (VL) utility similarity. The intervals [0.2, 0.4), [0.4, 0.6), [0.6, 0.8), and [0.8, 1] indicate synthetic images with progressively higher levels of utility similarity, *i.e.*, VL, Low (L), Moderate (M), High (H) and Very-high (VH), respectively.

The classification performance achieved using datasets of different sizes (60%, 80%, 100% of the training data), comprising synthetic images belonging to different IUS score intervals for training, are illustrated in Fig. 2. It can be noticed that synthetic image sets with IUS scores falling within the interval VH yield the best classification performance. This result was consistent for synthetic samples deriving from both *Generators I* and *II*. Moreover, it can be observed that synthetic training image sets deriving from *Generator-II* achieved a lower classification performance (Fig. 2(b)) than the respective sets deriving from *Generator-I* (Fig. 2(a)), over all utility levels. This could be attributed to the information abstraction via latent representations within VAE architectures [3]. Furthermore, similar results were obtained even with the smaller training subsets (Fig. 2). Based on these results, in the rest of this paper two IUS score categories are considered: VH and not (¬) VH, *i.e.*, $\mathcal{U} \in [-1, 0.8)$.

To assess the correlation between the IUS measure and image quality, we compared the FID scores of synthetic images belonging to different IUS levels (Table III). It can be noticed that FID provides better (↓) scores when applied on VH images, indicating a high correlation with IUS scoring scale. This shows that the use of IUS for the selection of higher utility datasets also results in datasets of higher fidelity. Furthermore, we investigated the correlation between the IUS measure and the IS score, which measures image diversity while exhibiting a limited expressivity in the assessment of synthetic medical image fidelity. This can be attributed to its reliance on ImageNet pretrained class probabilities (rather than feature distribution statistics as is the case of FID), which are not representative for medical images [13], [24]. Regarding diversity, as expressed by IS, it can be noticed that it remains approximately unaffected over the different IUS levels without any systematic deviations.

## C. Impact Analysis

The impact of IUS as a measure assessing the utility similarity between synthetic and real medical images was evaluated in two phases. In the first phase classifiers were trained on synthetic images and tested on real images. The experiments aim to investigate if training the classifiers solely with synthetic images having IUS VH, results in better classification of real images than training the classifiers with randomly selected synthetic images, regardless of their IUS.

In the second phase the classifiers were trained with real datasets and then they were tasked to classify synthetic images. The experiments aim to investigate if synthetic images characterized by higher IUS better resemble the real ones, expecting that synthetic images resembling more the real ones will result in better classification performance.

In both phases, ResNet50 [15] and InceptionV3 [16], [40] implemented the classification tasks, using the parameters described in the previous section. In these phases, the synthetic image datasets utilized, were created so that they preserve the class distribution of the real datasets.

### 1) Improving real image classification using high utility synthetic training datasets

In the first evaluation phase, each synthetic dataset generated with *Generators I* and *II* is classified by a dedicated EPU-CNN model that has been trained on a respective real image dataset (Section III.A). For instance, a dermoscopic synthetic dataset is classified by an EPU-CNN model trained on ISIC. Then the EPU-CNN model is used to calculate $\mathcal{U}$ for each synthetic image. To generate a synthetic image dataset of very-high utility similarity (VH), images with IUS L score are discarded

TABLE IV
REAL ENDOSCOPIC IMAGE CLASSIFICATION RESULTS USING SYNTHETIC TRAINING DATA

| Data Generator | Utility | KID | | | | Kvasir - Capsule | | | | Kvasir v2 | | | |
|---|---|---|---|---|---|---|---|---|---|---|---|---|---|
| | | InceptionV3 | | ResNet50 | | InceptionV3 | | ResNet50 | | InceptionV3 | | ResNet50 | |
| | | Accuracy | AUC | Accuracy | AUC | Accuracy | AUC | Accuracy | AUC | Accuracy | AUC | Accuracy | AUC |
| *Generator-I* | - | 82.9 ± 3.4 | 90.4 ± 2.7 | 73.5 ± 1.4 | 83.9 ± 1.5 | 83.5 ± 1.3 | 88.3 ± 1.0 | 73.8 ± 2.7 | 79.2 ± 2.4 | 78.3 ± 1.7 | 84.2 ± 1.7 | 79.6 ± 1.7 | 84.4 ± 1.4 |
| *Generator-I* | *VH* | **87.0 ± 3.2** | 91.0 ± 1.6 | **87.8 ± 1.8** | 90.6 ± 2.0 | **85.3 ± 0.8** | **90.2 ± 0.5** | **81.9 ± 0.9** | **85.8 ± 1.4** | **82.7 ± 0.6** | 87.3 ± 1.1 | 83.4 ± 1.4 | **88.2 ± 1.0** |
| *Generator-I$_{80\%}$** | - | 82.1 ± 3.8 | 86.7 ± 3.4 | 69.0 ± 2.6 | 72.4 ± 2.2 | 84.9 ± 0.9 | 88.9 ± 0.9 | 74.2 ± 1.5 | 79.0 ± 1.4 | 80.8 ± 1.2 | 86.9 ± 1.0 | 80.5 ± 1.5 | 85.2 ± 1.1 |
| *Generator-I$_{80\%}$* | *VH* | 83.9 ± 2.3 | **91.1 ± 2.2** | 85.3 ± 3.3 | **90.8 ± 1.7** | 84.9 ± 1.3 | 89.6 ± 0.5 | 79.9 ± 2.3 | 84.5 ± 1.9 | 82.1 ± 2.1 | **88.4 ± 1.6** | **83.5 ± 0.9** | 87.8 ± 1.1 |
| *Generator-I$_{60\%}$* | - | 54.0 ± 0.2 | 65.5 ± 1.7 | 62.5 ± 1.2 | 67.1 ± 3.1 | **85.3 ± 0.8** | 90.1 ± 0.9 | 77.8 ± 0.9 | 82.0 ± 1.0 | 78.4 ± 1.8 | 83.9 ± 2.1 | 79.4 ± 1.5 | 85.7 ± 1.4 |
| *Generator-I$_{60\%}$* | *VH* | 77.6 ± 4.9 | 86.0 ± 3.4 | 87.2 ± 1.5 | 89.9 ± 1.1 | 82.0 ± 1.1 | 86.9 ± 1.1 | 81.0 ± 1.4 | 84.7 ± 1.8 | 80.0 ± 1.5 | 85.1 ± 1.7 | 81.0 ± 1.3 | 86.4 ± 0.9 |
| *Generator-II* | - | 87.0 ± 2.3 | 90.0 ± 2.4 | 80.9 ± 2.8 | 86.3 ± 2.6 | 68.7 ± 1.8 | 71.6 ± 1.9 | 66.9 ± 1.6 | 69.9 ± 1.6 | 75.9 ± 1.0 | 81.6 ± 0.9 | 75.5 ± 1.0 | 81.2 ± 0.6 |
| *Generator-II* | *VH* | 85.1 ± 1.4 | 91.7 ± 2.0 | **87.4 ± 1.9** | **91.4 ± 2.1** | **74.7 ± 1.3** | **79.0 ± 1.5** | **71.4 ± 1.5** | **73.4 ± 1.7** | **80.3 ± 1.5** | **85.7 ± 1.3** | **79.1 ± 1.2** | **83.5 ± 1.0** |
| *Generator-II$_{80\%}$* | - | 81.1 ± 1.3 | 86.9 ± 1.2 | 81.3 ± 1.7 | 86.3 ± 1.6 | 71.5 ± 1.2 | 74.6 ± 1.6 | 67.2 ± 1.6 | 67.3 ± 1.3 | 74.3 ± 1.8 | 80.2 ± 1.7 | 70.6 ± 0.3 | 77.2 ± 0.5 |
| *Generator-II$_{80\%}$* | *VH* | 84.9 ± 0.4 | 89.1 ± 1.5 | 84.3 ± 1.9 | 88.7 ± 0.8 | 72.5 ± 1.2 | 74.9 ± 1.4 | 68.9 ± 1.7 | 71.4 ± 1.8 | 79.3 ± 1.1 | 84.3 ± 1.2 | 73.8 ± 1.1 | 78.8 ± 0.4 |
| *Generator-II$_{60\%}$* | - | 54.0 ± 0.2 | 59.5 ± 1.5 | 54.0 ± 0.2 | 65.5 ± 1.2 | 68.1 ± 1.7 | 70.1 ± 1.9 | 65.5 ± 1.2 | 66.7 ± 1.4 | 73.3 ± 1.6 | 80.1 ± 1.3 | 69.9 ± 1.4 | 76.6 ± 0.6 |
| *Generator-II$_{60\%}$* | *VH* | **87.6 ± 2.3** | **92.0 ± 1.6** | 83.5 ± 2.7 | 87.2 ± 1.8 | 70.3 ± 1.9 | 72.2 ± 2.1 | 68.8 ± 2.1 | 70.3 ± 2.1 | 77.5 ± 1.0 | 83.8 ± 1.2 | 76.5 ± 1.4 | 80.4 ± 1.7 |

**Subscript 80% and 60% in the Data Generator indicates that the CNN training was performed using 80% and 60% of the initial training data, respectively.*

TABLE V
REAL DERMOSCOPIC IMAGE CLASSIFICATION RESULTS USING SYNTHETIC TRAINING DATA

| Data Generator | Utility | Melanoma vs Nevi | | | | Carcinoma vs Nevi | | | | Melanoma vs Carcinoma | | | |
|---|---|---|---|---|---|---|---|---|---|---|---|---|---|
| | | InceptionV3 | | ResNet50 | | InceptionV3 | | ResNet50 | | InceptionV3 | | ResNet50 | |
| | | Accuracy | AUC | Accuracy | AUC | Accuracy | AUC | Accuracy | AUC | Accuracy | AUC | Accuracy | AUC |
| *Generator-I* | - | 69.8 ± 1.8 | 74.9 ± 1.4 | 67.9 ± 0.9 | 72.5 ± 1.2 | 84.5 ± 0.9 | 89.0 ± 0.8 | 84.9 ± 0.3 | 89.4 ± 0.9 | 78.7 ± 1.5 | 83.8 ± 1.2 | 76.9 ± 0.7 | 82.2 ± 1.0 |
| *Generator-I* | *VH* | **72.2 ± 1.9** | **76.3 ± 1.5** | **71.9 ± 1.5** | 75.1 ± 1.1 | 84.0 ± 0.4 | 89.5 ± 0.2 | **85.6 ± 1.1** | **90.2 ± 0.6** | 78.6 ± 1.0 | **84.6 ± 0.7** | 77.4 ± 0.4 | 83.1 ± 0.5 |
| *Generator-I$_{80\%}$* | - | 70.0 ± 1.7 | 75.3 ± 1.7 | 68.9 ± 1.2 | 72.6 ± 1.4 | 84.7 ± 0.5 | 89.5 ± 0.4 | 84.2 ± 1.3 | 88.7 ± 1.1 | 75.5 ± 1.6 | 81.2 ± 1.3 | 76.3 ± 0.9 | 82.3 ± 0.9 |
| *Generator-I$_{80\%}$* | *VH* | 71.5 ± 0.9 | 76.0 ± 1.7 | 70.7 ± 1.4 | 73.9 ± 1.6 | **84.7 ± 0.9** | **90.0 ± 0.3** | 84.5 ± 0.8 | 89.7 ± 1.0 | 78.4 ± 0.9 | 83.3 ± 1.0 | 78.6 ± 1.1 | **83.9 ± 0.5** |
| *Generator-I$_{60\%}$* | - | 70.1 ± 2.3 | 74.3 ± 2.3 | 68.3 ± 1.4 | 72.1 ± 1.7 | 83.0 ± 1.1 | 87.7 ± 0.9 | 84.7 ± 0.4 | 88.5 ± 0.8 | 76.0 ± 1.3 | 81.2 ± 1.5 | 75.2 ± 1.8 | 80.7 ± 1.5 |
| *Generator-I$_{60\%}$* | *VH* | 70.8 ± 2.6 | 75.2 ± 2.3 | 71.5 ± 1.2 | **75.6 ± 1.8** | 83.5 ± 0.8 | 88.1 ± 0.6 | 84.7 ± 1.0 | 88.2 ± 0.5 | **78.7 ± 1.2** | 83.5 ± 1.2 | **78.8 ± 1.0** | 83.2 ± 0.7 |
| *Generator-II* | - | 56.6 ± 0.8 | 59.2 ± 1.2 | 54.4 ± 1.4 | 54.4 ± 1.4 | 60.5 ± 1.4 | 63.3 ± 1.1 | 60.5 ± 1.5 | 61.9 ± 1.8 | 53.4 ± 0.0 | 63.3 ± 0.8 | 53.4 ± 0.0 | 56.6 ± 0.8 |
| *Generator-II* | *VH* | 66.5 ± 0.9 | 71.3 ± 0.6 | **66.1 ± 1.2** | **70.7 ± 1.1** | **78.2 ± 1.1** | 83.4 ± 1.7 | 79.5 ± 0.7 | **85.0 ± 0.8** | 68.5 ± 1.0 | 74.8 ± 0.9 | **72.5 ± 1.1** | 76.2 ± 1.0 |
| *Generator-II$_{80\%}$* | - | 58.7 ± 1.3 | 62.5 ± 1.0 | 53.8 ± 0.8 | 55.2 ± 0.9 | 57.8 ± 1.8 | 60.4 ± 2.0 | 61.0 ± 1.6 | 64.8 ± 1.6 | 56.4 ± 0.9 | 58.6 ± 0.6 | 53.4 ± 0.0 | 57.5 ± 1.1 |
| *Generator-II$_{80\%}$* | *VH* | **68.5 ± 1.2** | **73.2 ± 0.8** | 62.5 ± 1.4 | 66.8 ± 1.2 | 74.7 ± 1.7 | 80.9 ± 1.4 | **80.8 ± 1.2** | 84.8 ± 0.8 | 72.2 ± 0.8 | 76.9 ± 0.8 | 70.6 ± 1.5 | **77.2 ± 1.4** |
| *Generator-II$_{60\%}$* | - | 58.1 ± 1.4 | 60.7 ± 2.2 | 53.7 ± 1.3 | 54.6 ± 2.3 | 53.2 ± 0.1 | 58.8 ± 1.1 | 58.1 ± 1.3 | 60.1 ± 1.9 | 53.4 ± 0.0 | 59.0 ± 0.5 | 53.4 ± 0.0 | 61.7 ± 0.7 |
| *Generator-II$_{60\%}$* | *VH* | 60.4 ± 1.2 | 63.3 ± 1.7 | 63.3 ± 1.0 | 66.1 ± 1.5 | 77.8 ± 1.3 | **84.0 ± 0.9** | 79.3 ± 0.5 | 83.4 ± 0.7 | **72.4 ± 1.2** | **78.0 ± 1.0** | 69.8 ± 0.7 | 76.3 ± 0.6 |

TABLE VI
REAL RETINAL IMAGE CLASSIFICATION RESULTS USING SYNTHETIC TRAINING DATA

| Data Generator | Utility | APTOS | | | | DeepDRiD | | | |
|---|---|---|---|---|---|---|---|---|---|
| | | InceptionV3 | | ResNet50 | | InceptionV3 | | ResNet50 | |
| | | Accuracy | AUC | Accuracy | AUC | Accuracy | AUC | Accuracy | AUC |
| *Generator-I* | - | **88.1 ± 1.5** | 91.6 ± 1.8 | 89.9 ± 1.1 | **93.6 ± 0.9** | 77.4 ± 2.6 | 82.6 ± 2.6 | 71.3 ± 2.7 | 73.9 ± 3.1 |
| *Generator-I* | *VH* | 86.9 ± 1.5 | **91.8 ± 1.5** | 86.9 ± 1.6 | 90.5 ± 1.2 | **81.8 ± 2.5** | 82.4 ± 2.6 | 77.4 ± 3.7 | 79.4 ± 3.5 |
| *Generator-I$_{80\%}$* | - | 87.6 ± 2.0 | **91.8 ± 1.5** | 89.3 ± 0.6 | 92.6 ± 0.9 | 79.4 ± 3.8 | 82.6 ± 3.3 | 71.4 ± 4.6 | 76.2 ± 3.5 |
| *Generator-I$_{80\%}$* | *VH* | 87.5 ± 1.8 | 90.7 ± 1.8 | 86.9 ± 1.7 | 90.2 ± 1.5 | 77.4 ± 4.4 | **83.7 ± 3.7** | **77.9 ± 3.1** | **79.4 ± 3.0** |
| *Generator-I$_{60\%}$* | - | 87.7 ± 1.0 | 91.5 ± 1.1 | **90.3 ± 1.7** | 92.7 ± 1.0 | 65.3 ± 3.7 | 70.6 ± 4.0 | 64.5 ± 4.1 | 71.7 ± 5.7 |
| *Generator-I$_{60\%}$* | *VH* | 86.0 ± 1.7 | 90.2 ± 0.6 | 87.0 ± 1.1 | 90.1 ± 1.2 | 77.3 ± 3.1 | 79.4 ± 3.8 | 71.6 ± 4.1 | 72.5 ± 4.7 |

and replaced with newly generated samples with IUS VH score. Two instances of the two CNN classifiers are trained separately with the VH set, and another control set of equal size is created by randomly sampling images from the original synthetic data, regardless of their IUS scoring results. Each trained instance of both networks is evaluated on real-image test sets. To further investigate the robustness of IUS-based synthetic image selection process, the training was repeated with 80% and 60% of each training set.

Tables IV–VI summarize the classification performance of the InceptionV3 and ResNet50 classifiers on real endoscopic (Table IV), dermoscopic (Table V) and retinal (Table VI) datasets, after being trained on respective synthetic samples, using samples not selected based on their IUS score (denoted with a dash “–”), and comparatively, using samples with IUS score VH. The results indicate that the usage of synthetic training samples with IUS VH can lead to performance improvements. Specifically, for the endoscopic data, the training sets solely comprising samples with IUS VH scores, showed improved classification performance, compared to the randomly selected training sets. In the case of the KID dataset, using training images with IUS VH deriving from *Generator-I*, InceptionV3 increased its AUC from 83.9±2.0% to 90.6±2.0%, yielding a relative improvement of 8.0%. In the case of the Kvasir-Capsule dataset, the IUS VH images from *Generator-II* increased the AUC of InceptionV3 from 71.6±1.9% to 79.0±1.5%, yielding a relative improvement of 10.3%. In the case of the Kvasir v2 dataset, the IUS VH images from *Generator-II*, increase the AUC of InceptionV3 from 81.6±0.9% to 85.7±1.3%, yielding a relative improvement of 5.0%. It can be observed that similar improvement trends were observed even when the training data were reduced to 80% and 60%. Notably, for KID dataset, the IUS VH images deriving from *Generator-II* resulted in relative AUC improvement up to 54.6%, when training InceptionV3 with only 60% of the initial training set.

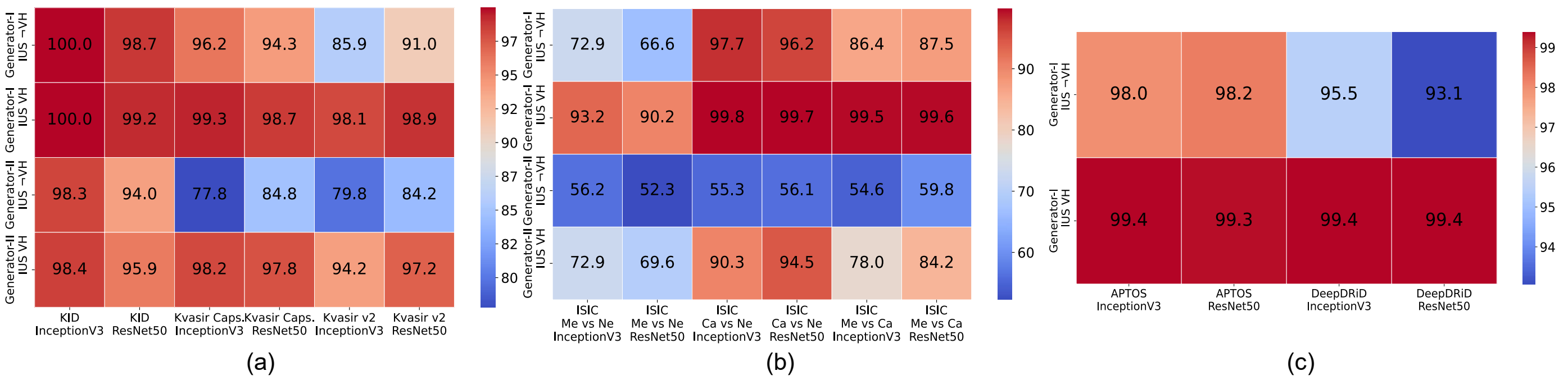

Fig. 3. Classification performance obtained after training on real, and testing on synthetic image subsets with IUS VH and IUS ¬VH. Results for (a) endoscopic, (b) dermoscopic and (c) retinal datasets.

Table V shows that classifiers trained with synthetic dermoscopic sets consisting solely of IUS VH samples consistently lead to enhanced performance across all experiments. The IUS VH training sets deriving from *Generator-II* yielded the most notable improvements. For instance, the ResNet50 classifier achieved relative AUC improvements of 30.0% (from 54.4±1.4% to 70.7±1.1%) classifying *Me vs Ne*, 37.3% (from 61.9±1.8% to 85.0±0.8%) classifying *Ca vs Ne*, and 34.6% (from 56.6±0.8% to 76.2±1.0%) classifying *Me vs Ca*. Notably, these trends were maintained even when the size of training sets were reduced. Using 60% of the IUS VH images, best relative improvements in AUC were observed in the classification of *Ca vs Ne*. Specifically, the performance of IncpetionV3 increased by 42.9%, and the performance of ResNet50 by 38.8. Similar improvements were observed also in terms of accuracy.

Regarding the retinal datasets (Table VI), the synthetic images generated based on the APTOS dataset with IUS VH the best improvement achieved by the classifiers using the full dataset in terms of AUC was 0.2%. Regarding the synthetic images generated based on the DeepDRiD dataset, the relative improvement in terms of AUC was 7.4% with ResNet50. Moreover, the IUS VH images led in gains in classification performance even with less training data. Specifically, with 60% of training data, InceptionV3 achieved relative AUC and accuracy improvements of 12.5% and 18.4%, respectively. It should be noted that *Generator-II* was not included in the evaluation of retinal datasets because it resulted in exploding KL divergence and failure in image synthesis.

Overall, it can be observed that the performance gains obtained using IUS for the selection of synthetic training image subsets vary across the different imaging data modalities considered in this study. For example, the performance gain obtained using IUS to select a useful training subset in the case of the retinal fundus imaging was smaller than the respective gains obtained in the case of other imaging modalities. This could be attributed to the subtle anatomical structures contained in the real retinal fundus images, *e.g.*, retinal vessel and arterial trunks, distinct optic disc shape, macula color pigmentation, which are challenging to be fully reproduced in synthetic images.

### 2) Improving the resemblance with real images using high utility synthetic images

In this evaluation phase, the classifiers were trained with real data and their ability to accurately classify synthetic samples with IUS VH and ¬VH was assessed. The experimental procedure involved 3 steps: a) The IUS measure was used to organize the synthetic samples into two subsets: one consisting solely of images with IUS VH, and one consisting of images with IUS ¬VH. b) The classifiers were then trained with real images. c) The trained classifiers were used to classify the images of each synthetic subset, considering that a higher classification result signifies improved resemblance with the real images, based on image features captured by the classifiers.

The classification results obtained in terms of AUC scores are summarized in Fig. 3 for the endoscopic, dermoscopic, and retinal datasets, respectively. It can be noticed that synthetic images with IUS VH provided improved AUC scores compared to their ¬VH counterparts across all modalities. Specifically, regarding the endoscopic datasets (Fig.3(a)) all classifiers provided consistently higher AUC scores when applied on the synthetic samples with IUS VH. ResNet50 resulted in an improved AUC using synthetic images with IUS VH from *Generator-II* based on the KID dataset. The improvement was 1.9% compared to ¬VH (VH: 95.9% vs ¬VH: 84.0%). The highest AUC improvements for Kvasir-Capsule and Kvasir-v2 datasets were observed using Inception-V3 and the same generator, with gains of 20.3% (VH: 98.2% vs ¬VH:77.8%) and 14.4% (VH:94.2% vs ¬VH:79.8%), respectively. In the case of dermoscopic images (Fig.3(b)), the subsets with IUS VH deriving from *Generator-I* and *II* outperformed the ¬VH subsets in all evaluation scenarios providing performance improvements in terms of AUC up to 38.4% (Ca *vs* Ne, *Generator-II,* ResNet50). It can be noticed that the synthetic images with IUS ¬VH generated with *Generator-II* resulted in AUC scores of ~50% (in Me *vs* Ne, ResNet50 and in Me *vs* Ca, *InceptionV3*), indicating random performance. In the case of the retinal datasets (Fig.3(c)) the synthetic subsets with IUS VH demonstrated improved AUC compared to subsets with IUS ¬VH. Notably, for the synthetic images generated based on the DeepDRiD dataset the classification of images with IUS VH by ResNet50 was improved 6.3% compared to the ¬VH images (VH:99.4% vs ¬VH:93.1%). Overall, the classification results show that regardless of the medical imaging modality considered, the synthetic images with IUS VH are classified more accurately than those with IUS ¬VH, indicating that exhibit a higher resemblance with the respective real images used to train the classifiers.

## D. Qualitative Results

A qualitative analysis of the IUS-based assessment for both real and synthetic images is presented in Figs. 4-7. Figure 4 shows representative real images with IUS VH. The endoscopic images (Fig. 4(a)-(c)) illustrate tissues with distinct anatomical structures and lesions having expected color and texture properties. The dermoscopic images (Fig. 4(d)) adhere to the criteria of the ABCD rule (Asymmetry, Border irregularities, Color variations inside, and inhomogeneous Dermoscopic structures) [41]. Retinal fundus images (Fig. 4(e)-(f)) clearly exhibit eye fundus structures and portraying findings, such as exudate fluids, conforming to diabetic retinopathy (DR) conditions. Figure 5 illustrates representative real images with IUS ¬VH. It can be observed that the endoscopic images (Fig.5(a)-(c)) exhibit limited anatomical detail, containing lesions with atypical morphology or unnatural positioning in the epithelium and occasionally the presence of artifacts is visible (Fig. 5(a), second row); demonstrating the IUS capacity to identify such images and classify them as of lower utility.

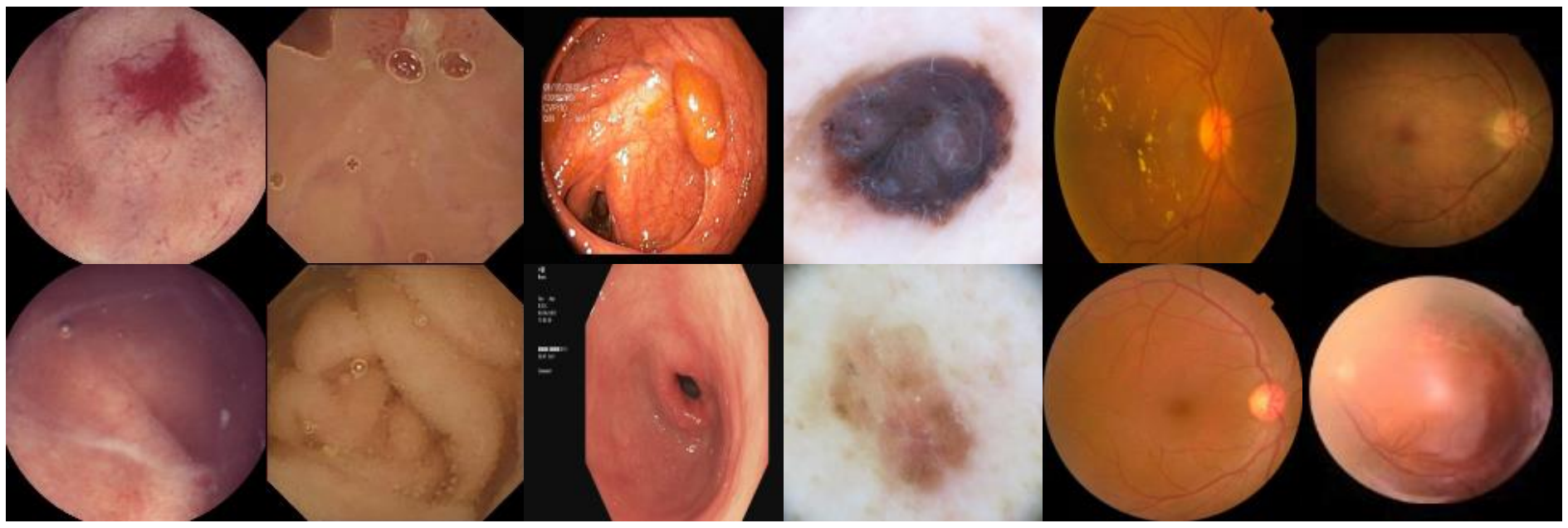

(a) (b) (c) (d) (e) (f)

Fig. 4. Real images with IUS VH, illustrating abnormal (row 1) and normal (row 2) conditions from various datasets (a) KID, (b) Kvasir-Capsule, (c) Kvasir-v2, (d) ISIC, (e) APTOS, (f) DeepDRiD.

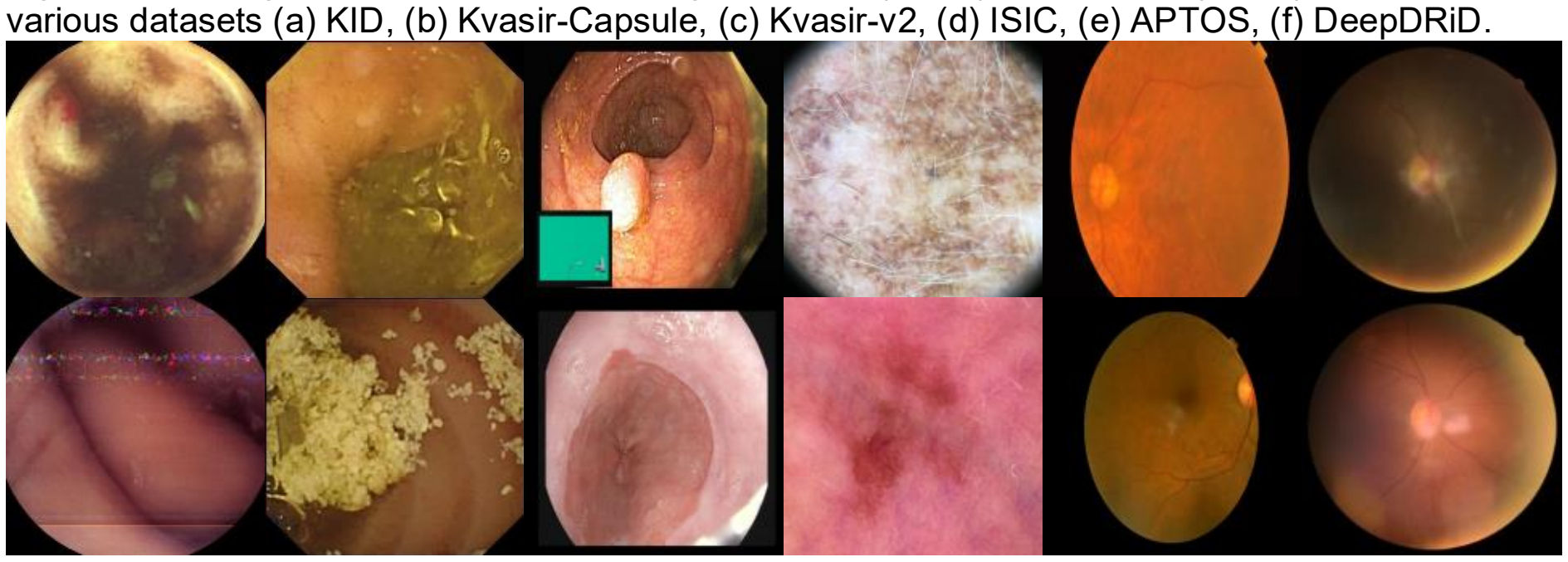

(a) (b) (c) (d) (e) (f)

Fig. 5. Real images with IUS ¬VH, illustrating abnormal (row 1) and normal (row 2) conditions from various datasets. (a) KID, (b) Kvasir-Capsule, (c) Kvasir-v2, (d) ISIC, (e) APTOS, (f) DeepDRiD.

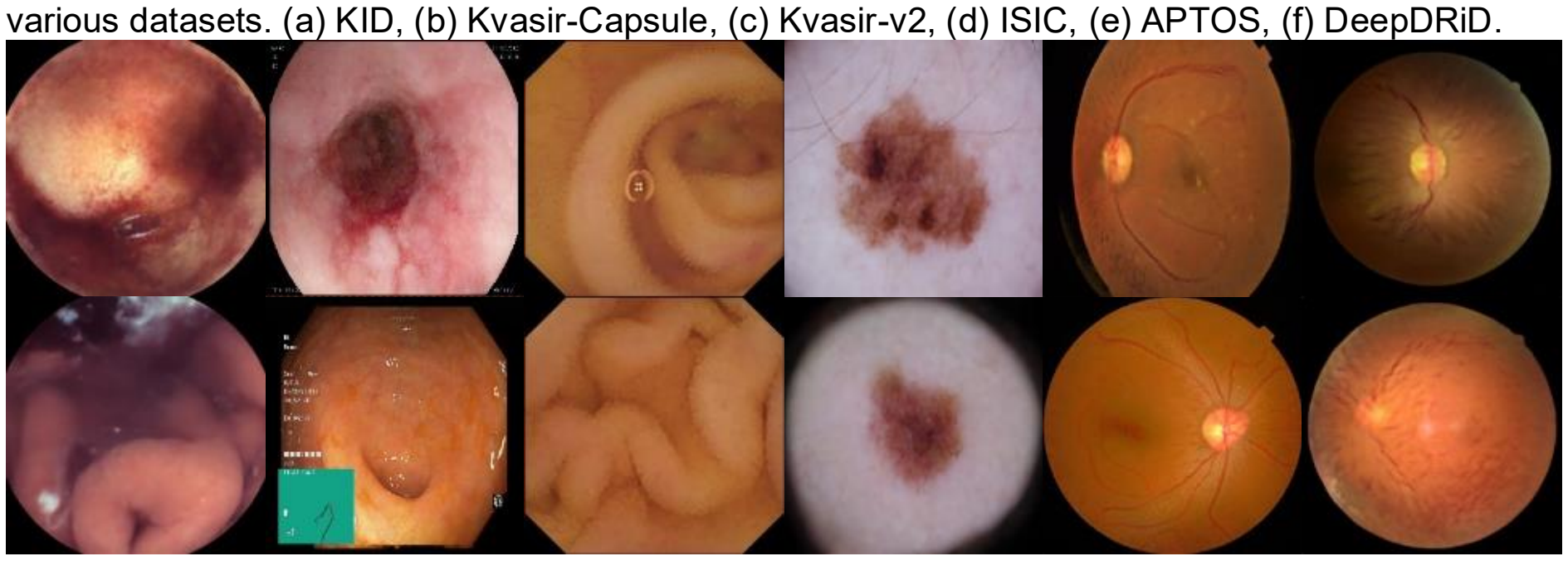

(a) (b) (c) (d) (e) (f)

Fig. 6. Synthetic abnormal (row 1) and normal (row 2) images with IUS VH from various datasets. (a) KID, (b) Kvasir v2, (c) Kvasir-Capsule, (d) ISIC, (e) APTOS, and (f) DeepDRiD.

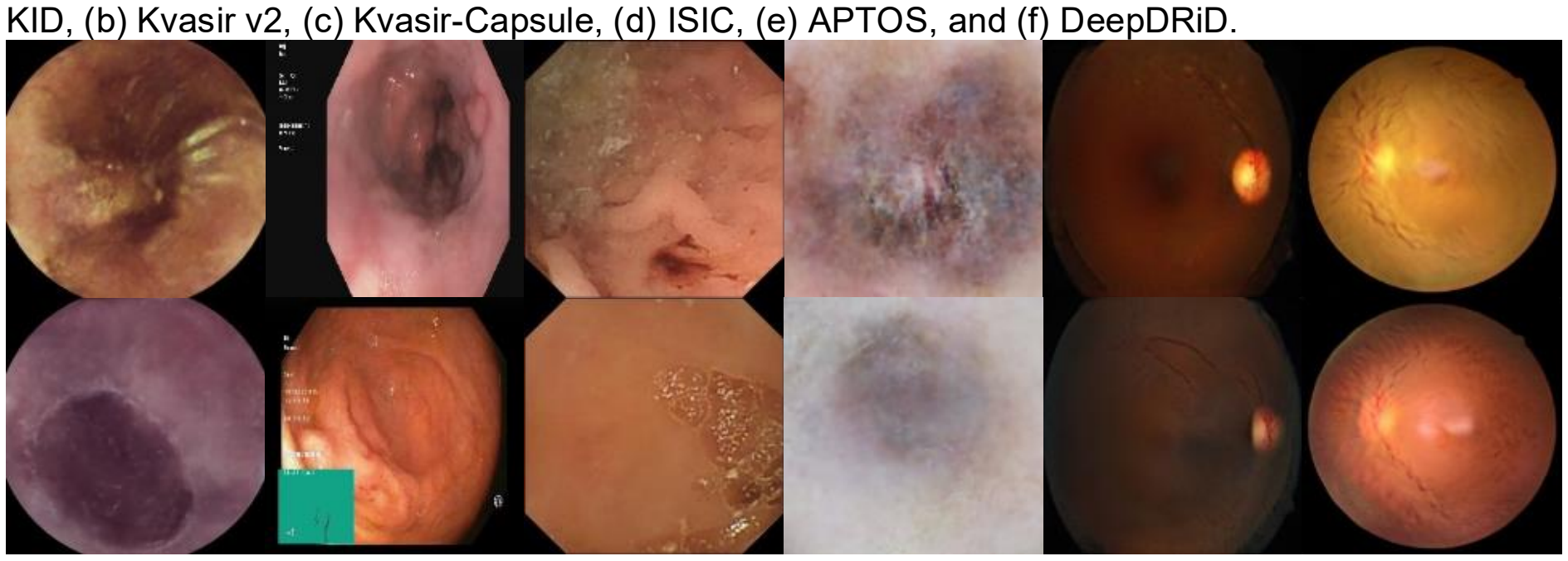

(a) (b) (c) (d) (e) (f)

Fig. 7. Synthetic abnormal (row 1) and normal (row 2) images with IUS ¬VH from various datasets. (a) KID, (b) Kvasir v2, (c) Kvasir-Capsule, (d) ISIC, (e) APTOS, and (f) DeepDRiD.

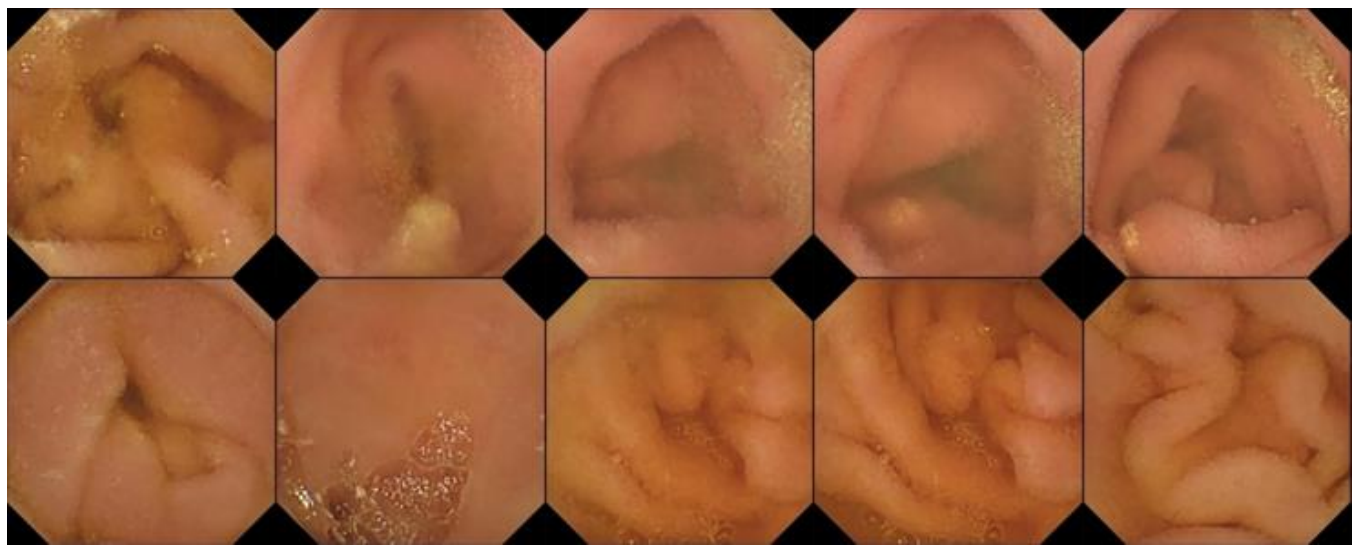

(a) (b) (c) (d) (e)

Fig. 8. Representative synthetic abnormal (row 1) and normal (row 2) endoscopic images generated based on KID, with IUS: (a) VL (b) L, (c) M (d) H (e) VH.

The dermoscopic images (Fig. 5(d)) have pixelated textures, color discontinuities and structures that misrepresent the clinical characteristics of skin conditions. Retinal images (Fig. 5(e)-(f)) suffer from darkness and haziness that disturb the observation of fundus anatomy. Figures 6 and 7 illustrate respective results with those of Fig. 4 and Fig. 5 but with synthetic images. It can be noticed that the endoscopic images (Fig.6(a)-(c)), illustrate tissues that effectively reproduce the color and texture of mucosa as well as they reflect the appearance of pathological conditions. The dermoscopic images (Fig. 6(d)) conform to the diagnostic criteria of the ABCD rule, effectively capturing the clinical characteristics associated with skin conditions. Retinal fundus images (Fig. 6(e)-(f)) replicate optic disc color, optic cup shape, macula appearance and include DR-related findings consistent to the retinal pathology. Synthetic images with IUS ¬VH are illustrated in Fig. 7. The endoscopic images (Fig.7(a)-(c)) exhibit blurriness and haziness while they display inconsistencies in color and texture features. These issues are particularly evident in cases of pathological conditions. The dermoscopic images (Fig. 7(d)) are also characterized by haziness. Additionally, the dermoscopic findings are illustrated with unanticipated color and texture variations or deformations in their structure which tend to distort the clinical features of skin lesions. The retinal fundus images (Fig. 7(e)(f)) exhibit dark or uneven illumination with unexpected optic cup shape, inconsistent choroidal texture and occasional pixelation.

An indicative visual example that shows how IUS relates with synthetic image content is illustrated for endoscopic images in Fig. 8. It can be observed that images with higher IUS scores better represent the appearance of endoscopic findings and display tissues with more informative detail. As the IUS score progressively increases, there is a noticeable improvement in clarity and representation of features in the synthetic images.

## E. Interpretation of Synthetic Image Utility

The IUS measure is based on the EPU-CNN framework which is inherently interpretable, considering perceptual concepts (in this paper color and texture). Thus, the results of IUS are expected to be interpretable upon these concepts. This has been qualitatively reflected to the results of the previous section, but it can also be quantified by the feature contribution profiles provided as interpretations by the EPU-CNN (section III.B). Figure 9 illustrates representative interpretation examples obtained for abnormal (Fig. 9(a-c)) and normal (Fig. 9(d-f)) images with IUS VH per dataset (in row 2), along with counterexamples, *i.e.*, images with IUS ¬VH (in row 3).

It can be observed that VH synthetic images follow the trends (ascending or descending) of the baseline feature contribution profile, *i.e.*, $C_b$ (Fig. 9, row 1). Conversely, the feature profile of images with IUS ¬VH as estimated by the interpretations of the EPU-CNN model deviates strongly from the baseline feature contribution profile $C_b$. Since IUS assesses the utility of synthetic images based on their feature contribution profiles (Section III.B), one can intuitively interpret why the IUS of an image is either ¬VH or VH w.r.t perceptual features of color and texture. More specifically, when an image with IUS ¬VH deviates from the baseline profile in specific feature components, such as C-F; the deviations imply visual flaws in the image. The endoscopic images with IUS ¬VH (Fig. 9(a), (b), (d), row 3) demonstrate blurriness (Fig. 9(a)-(b)), irregular contours (Fig. 9(b), (d)), and visual artifacts (Fig. 9(a)-(b), (d)). Accordingly, the PFMs related to these image properties demonstrate an opposite trend compared to the baseline contribution profile. The same holds across the dermoscopic and retinal cases. The dermoscopic images with IUS ¬VH (Fig. 9(e), row 3) are opposite to the baseline color (B-Y) and texture (C-F) contributions. These deviations can be attributed to noisy hue and texture. Both the retinal images with IUS ¬VH (Fig. 9(c),(f), row 3) lack fine texture and display color shifts. Notably, the image with IUS ¬VH in Fig. 9(c) exhibits a complete inversion of its feature contribution profile, which can be explained considering the absence of important anatomical features, *e.g.*, vessels, optic cup and its flat texture.

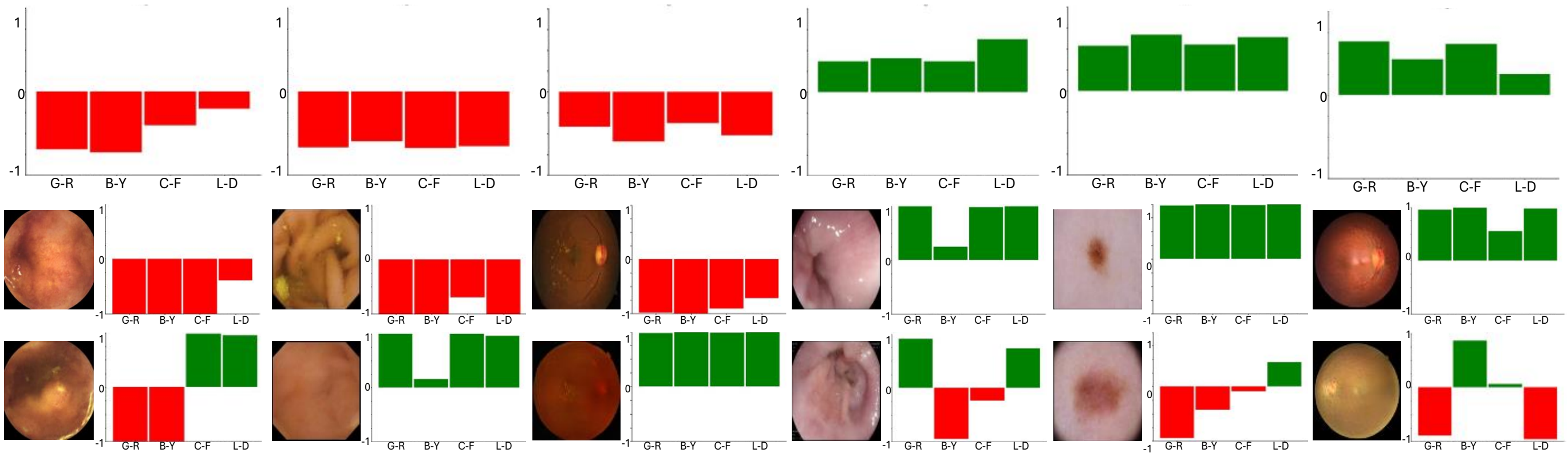

(a) (b) (c) (d) (e) (f)

Fig. 9. Representative examples of baseline feature contribution profiles $C_b$ (row 1) for various abnormal (a-c; red color) and normal (d-f; green color) classes of IUS VH (row 2) and IUS ¬VH (row 3) from various datasets. (a) KID, (b) Kvasir v2, (c) Kvasir-Capsule, (d) ISIC, (e) APTOS and (f) DeepDRiD.

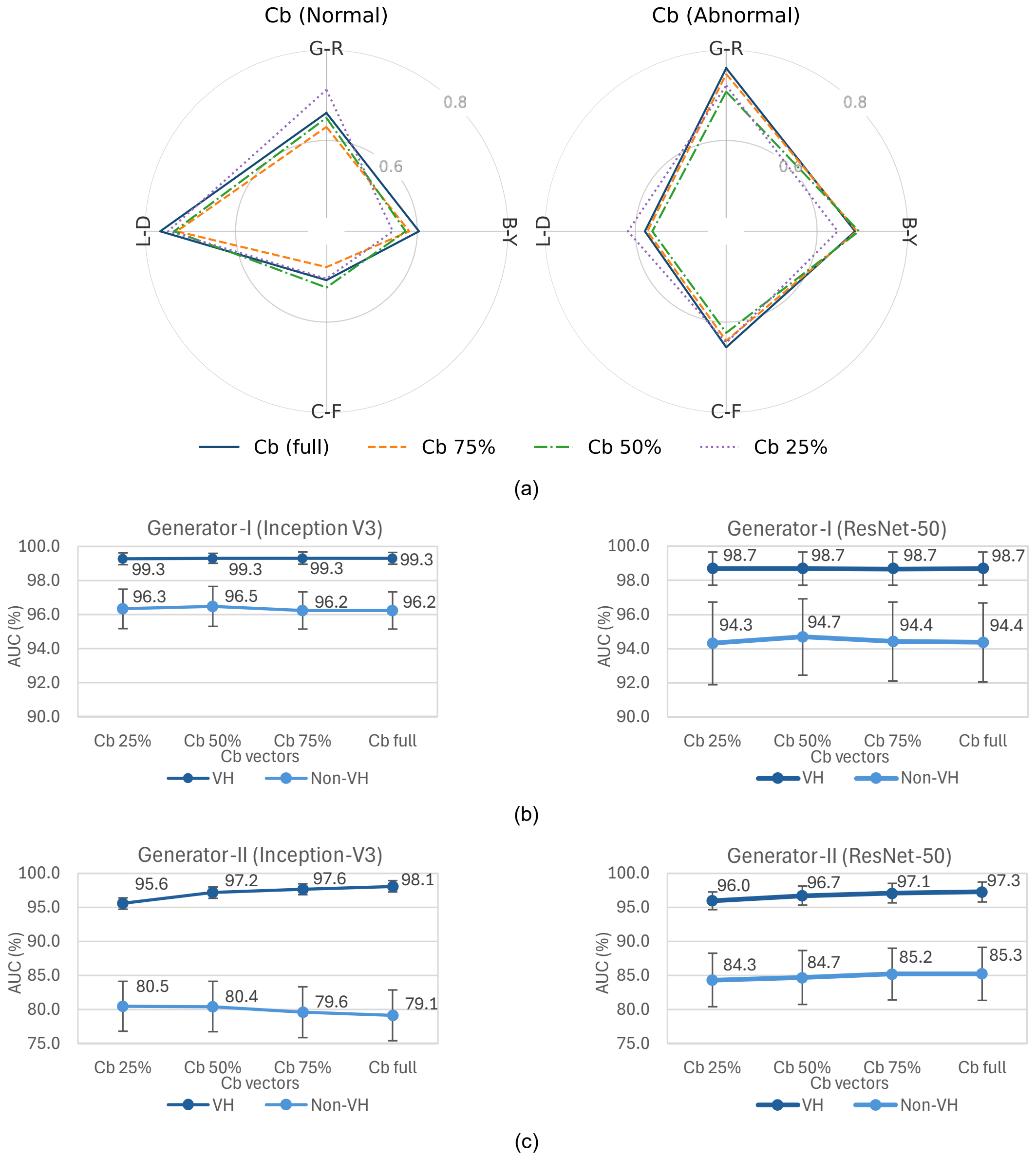

Fig. 10. Sensitivity analysis results. (a) Radar plot of vectors $C_b^{Full}$, $C_b^{75\%}$, $C_b^{50\%}$ and $C_b^{25\%}$ representing different configurations of baseline feature contribution profiles. (b)-(c) Classification performance obtained by training on real images and testing on synthetic images with IUS VH or ¬ VH using *Generators I* and *II*.

## F. IUS Sensitivity Analysis

To assess the robustness of IUS as a utility measure, its sensitivity was investigated with respect to the calculation of the baseline feature contribution profiles and the respective interpretations obtained.

### 1) Baseline profile sensitivity

An ablation study was performed to investigate the degree to which the number of real image feature contribution profiles $\boldsymbol{C}_i$, used to calculate the baseline feature contribution profile $\boldsymbol{C}_b$ (Eq. (6)), influences the evaluation of IUS. We consider $\boldsymbol{C}_b^{Full}$, representing such a baseline profile calculated using all the available profiles $\boldsymbol{C}_i$ from the test set. Additional baseline vectors were also calculated, namely $\boldsymbol{C}_b^{25\%}$, $\boldsymbol{C}_b^{50\%}$ and $\boldsymbol{C}_b^{75\%}$, using random 25%, 50% and 75% subsets of the test set, respectively. Figure 10(a) illustrates these vectors with different colors in a radar plot, where the values of its four components, namely G-R, B-Y, C-F, and L-D components, are depicted in four respective axes. It can be noticed that both in the case of normal and abnormal images, the vectors exhibit minimal differences between them, even with smaller size subsets used for their computation. This suggests that the baseline contribution profiles are quite tolerant to the number of real image contribution profiles used for their calculation. Subsequently, the various vectors $\boldsymbol{C}_b$ calculated for this ablation study, were used to evaluate the IUS measure (i.e., VH or ¬VH) of the synthetic images obtained from Generators I and II. The images with IUS VH and ¬VH were assessed following the experimental procedure described in Section III.C.2, i.e., classifiers trained on real data were separately tested on synthetic subsets consisting of images either with IUS VH or VH and ¬VH. The results confirm that the classification performance obtained using these two subsets is only limitedly affected across all the tested configurations of $\boldsymbol{C_b}$, for both Generator-I (Fig. 10(b)) and Generator-II (Fig. 10(c)).

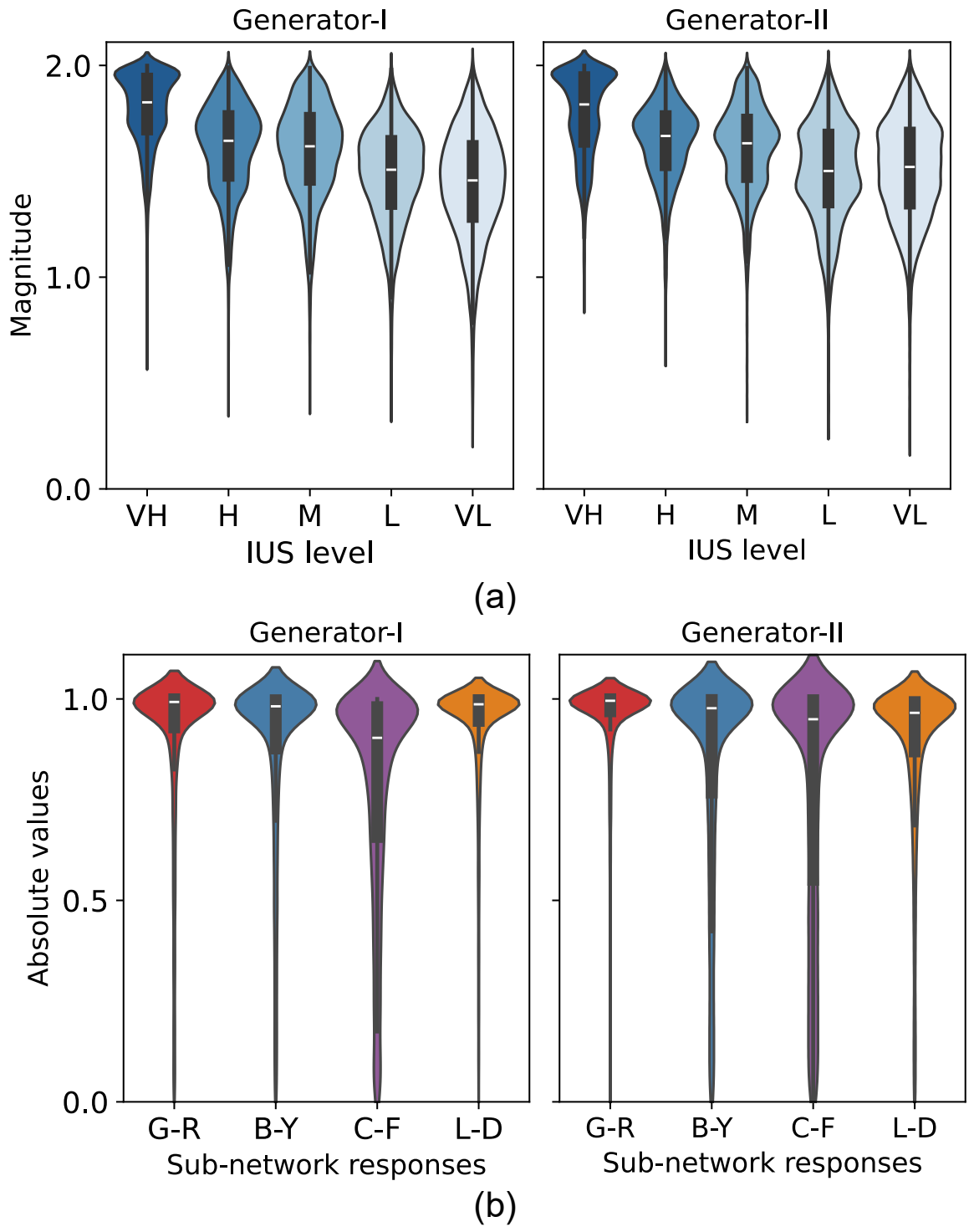

Fig. 11. Violin plots showing the distribution of magnitude for EPU-CNN (a) interpretation vectors across IUS levels (VH: very high, H: high, M: moderate, L: low, VL: very low) from Generator-I and -II synthetic images. (b) $f_i$ responses for IUS VH Generator-I and -II images, of G-R, B-Y, C-F and L-D PFMs. Wider violin sections represent higher densities of the respective magnitude values.

### 2) Interpretation Sensitivity

IUS relies on cosine similarity, which mainly considers the directionality of the vectors being compared. Therefore, when assessing colinear vectors of uniformly low magnitude as denoted by feature contribution profiles $\tilde{\boldsymbol{C}}$ (interpretation vectors), obtained from the synthetic images, may lead to less intuitive interpretations. To investigate the degree to which such cases may occur, a deeper analysis of the interpretation vectors was performed to gain further insights regarding their sensitivity. Figure 11(a) illustrates the distribution of the magnitudes of the interpretation vectors over a range of IUS levels. In this plot, each violin corresponds to a different utility level. Wider violin sections represent higher densities of the respective magnitude values. The y-axis range extends up to 2 since each of the four sub-networks yielding an output within the range of [–1, 1], as described in Section III.A, results in a maximum possible vector magnitude of 2. It can be noticed that overall, EPU-CNN tends to yield high magnitude responses for all utility levels. The interpretation vectors obtained from synthetic images with IUS VH, either generated from Generator-I or II, exhibit a skewed distribution towards the higher magnitude values. This indicates that most vector components, each corresponding to an EPU-CNN sub-network response, yield strong activations. In contrast, for synthetic images with IUS ¬VH, the interpretation vectors tend to have lower magnitude responses, exhibiting lower medians (white dots located in the center of the inner bars of Fig. 11(a)).

Figure 11(b) illustrates the distribution of the absolute values of the interpretation vector components obtained only from synthetic images with IUS VH. Each component, *i.e.*, G-R, B-Y, L-D and C-F, represents the respective EPU-CNN sub-network responses, falling within the interval of [-1, 1]. It can be noticed that the synthetic images with IUS VH have high magnitude responses over all sub-networks, either they originate from *Generator*-I or from *Generator*-II. This indicates that images with IUS VH contain features that strongly activate the respective perceptual image components.

The interpretation vectors were further analyzed, with respect to the probability of all their components being simultaneously under specific thresholds. To this end, thresholds from 0.0 to 1.0, with a step of 0.1, were considered. More specifically, for synthetic images originating from *Generator*-I, all the interpretation vectors had component values that were simultaneously above threshold 0.5, with a probability of 0.1% to provide responses equal to or below 0.5. For synthetic images originating from *Generator*-II, all the interpretation vectors had component values that were simultaneously above 0.7, with a probability of 0.0001% to provide responses equal to, or below 0.7. Notably, magnitude responses below these thresholds, *i.e.*, under 0.5 and 0.7, were never observed simultaneously. Therefore, uniformly low magnitude components in the interpretation vectors can be considered as an extremely rare case in the EPU-CNN formulation, which is designed to coordinate complementary feature interactions across sub-networks (Section III.A); thus, assessing the directionality of the interpretation vectors is more meaningful within the IUS framework.

## G. Expanding IUS evaluation in grayscale medical imaging modalities

Aiming to investigate the generalization of IUS measure in different domains, an IUS-based synthetic image utility evaluation was performed on grayscale medical datasets. The EPU-CNN model was adapted to grayscale settings as described in Section III.A, *i.e.,* the L-D and C-F texture PFMs were retained, whereas the color PFMs were replaced by two complementary PFMs corresponding to the lower and higher intensity ranges (since different grayscale intensities usually signify different tissues or findings), as proposed in [20]. BAND-I represents the PFM depicting intensities that belong in the first half of the grayscale, and BAND-II represents the PFM depicting intensities belonging in the second half of the grayscale. Following the training of the EPU-CNN model, the IUS measure was used according to Section IV.B to evaluate the utility of grayscale synthetic medical datasets from *Generator-I* and *II*. Additionally, a diffusion-based image generation method utilizing a Latent Diffusion Model (LDM) (*Generator-III*) [42] was included, considering the broad applicability of LDMs on grayscale modalities [1]. The evaluation was conducted on ultrasound (BreastMNIST) and X-ray (PneumoniaMNIST) benchmark datasets, both included in the MedMNIST collection [35]. The synthetic images with IUS

TABLE VI
REAL IMAGE CLASSIFICATION RESULTS USING SYNTHETIC TRAINING DATA FROM VARIOUS GRAYSCALE IMAGING MODALITIES

| Data Generator | Utility | *BreastMNIST* | | | | *PneumoniaMNIST* | | | |
|---|---|---|---|---|---|---|---|---|---|
| | | *InceptionV3* | | *ResNet50* | | *InceptionV3* | | *ResNet50* | |
| | | Accuracy | AUC | Accuracy | AUC | Accuracy | AUC | Accuracy | AUC |
| *Generator-I* | - | 80.5 ± 4.3 | 88.5 ± 2.3 | 74.6 ± 1.1 | 83.6 ± 2.5 | 91.9 ± 3.3 | 97.0 ± 2.3 | **88.6 ± 2.6** | 94.9 ± 2.0 |
| *Generator-I* | *VH* | **86.2 ± 8.0** | **91.3 ± 5.2** | **79.9 ± 5.0** | 86.5 ± 5.4 | 91.7 ± 3.5 | 96.9 ± 2.4 | 88.0 ± 3.9 | 94.9 ± 3.0 |
| *Generator-I$_{80\%}$* | - | 81.2 ± 3.8 | 86.7 ± 3.2 | 78.4 ± 6.0 | 83.2 ± 3.9 | 92.3 ± 3.0 | 97.0 ± 2.5 | 86.5 ± 4.0 | 94.5 ± 2.3 |
| *Generator-I$_{80\%}$* | *VH* | 86.0 ± 8.0 | 89.3 ± 6.2 | 79.2 ± 5.5 | **87.0 ± 6.7** | 91.6 ± 3.3 | 96.7 ± 2.4 | 88.6 ± 4.2 | **95.0 ± 3.3** |
| *Generator-I$_{60\%}$* | - | 73.5 ± 1.5 | 80.0 ± 5.7 | 73.6 ± 0.8 | 80.3 ± 1.6 | **92.4 ± 3.6** | **97.0 ± 2.1** | 85.5 ± 3.5 | 93.9 ± 2.7 |
| *Generator-I$_{60\%}$* | *VH* | 73.6 ± 0.8 | 80.3 ± 3.2 | 79.5 ± 4.6 | 86.2 ± 3.7 | 92.1 ± 3.5 | 97.0 ± 2.5 | 87.7 ± 4.8 | 94.5 ± 3.2 |
| *Generator-II* | - | 73.6 ± 0.8 | 82.0 ± 5.0 | 74.4 ± 6.9 | 80.7 ± 6.1 | 84.0 ± 1.5 | 87.0 ± 2.3 | 83.4 ± 3.9 | **90.0 ± 1.1** |
| *Generator-II* | *VH* | **80.2 ± 4.9** | **89.5 ± 5.2** | 79.7 ± 4.6 | 87.2 ± 6.7 | 84.1 ± 1.8 | 88.3 ± 1.7 | 82.9 ± 2.9 | 89.8 ± 0.6 |
| *Generator-II$_{80\%}$* | - | 73.6 ± 0.8 | 80.1 ± 5.0 | **80.8 ± 4.0** | 86.8 ± 5.0 | 83.6 ± 2.7 | 89.6 ± 1.6 | 76.5 ± 4.9 | 89.0 ± 3.2 |
| *Generator-II$_{80\%}$* | *VH* | 79.5 ± 5.0 | 88.7 ± 5.6 | 78.6 ± 6.0 | **88.3 ± 5.8** | 83.8 ± 3.0 | 89.7 ± 2.0 | 77.1 ± 5.1 | 89.6 ± 3.4 |
| *Generator-II$_{60\%}$* | - | 73.6 ± 0.8 | 77.0 ± 0.7 | 77.2 ± 2.5 | 82.9 ± 1.5 | **84.5 ± 3.0** | **90.0 ± 1.9** | **83.7 ± 3.7** | 89.8 ± 0.9 |
| *Generator-II$_{60\%}$* | *VH* | **80.2 ± 4.9** | 88.6 ± 3.9 | 73.6 ± 0.8 | 78.5 ± 1.9 | 84.0 ± 3.2 | 89.8 ± 2.0 | 82.7 ± 3.1 | 89.7 ± 0.8 |
| *Generator-III* | - | 69.5 ± 4.9 | 82.3 ± 3.6 | 70.0 ± 3.5 | 74.9 ± 2.6 | 84.9 ± 1.4 | **92.0 ± 1.0** | 83.9 ± 4.3 | 91.0 ± 2.9 |
| *Generator-III* | *VH* | 79.9 ± 4.5 | 87.7 ± 6.5 | 75.2 ± 1.7 | 82.0 ± 2.5 | 82.0 ± 2.4 | 88.0 ± 1.7 | 83.4 ± 3.5 | 91.0 ± 2.6 |
| *Generator-III$_{80\%}$* | - | 73.7 ± 0.8 | 81.6 ± 3.7 | 67.0 ± 3.3 | 75.0 ± 3.0 | 85.0 ± 1.4 | 91.6 ± 2.0 | 84.2 ± 4.9 | 91.3 ± 2.7 |
| *Generator-III$_{80\%}$* | *VH* | **82.2 ± 3.4** | 87.8 ± 6.3 | **82.3 ± 4.3** | **86.2 ± 5.4** | 81.1 ± 2.7 | 88.1 ± 1.3 | **84.4 ± 3.9** | **91.4 ± 2.8** |
| *Generator-III$_{60\%}$* | - | 73.7 ± 0.8 | 79.7 ± 4.4 | 73.6 ± 0.8 | 71.6 ± 2.0 | **86.2 ± 0.6** | 91.8 ± 1.0 | 83.4 ± 5.0 | 91.2 ± 3.2 |
| *Generator-III$_{60\%}$* | *VH* | 80.7 ± 4.1 | **88.0 ± 6.2** | 73.7 ± 0.8 | 72.8 ± 2.4 | 82.6 ± 2.0 | 87.9 ± 1.7 | 82.0 ± 4.9 | 91.0 ± 3.0 |

**Subscript 80% and 60% in the Data Generator indicates that the CNN training was performed using 80% and 60% of the initial training data, respectively.*

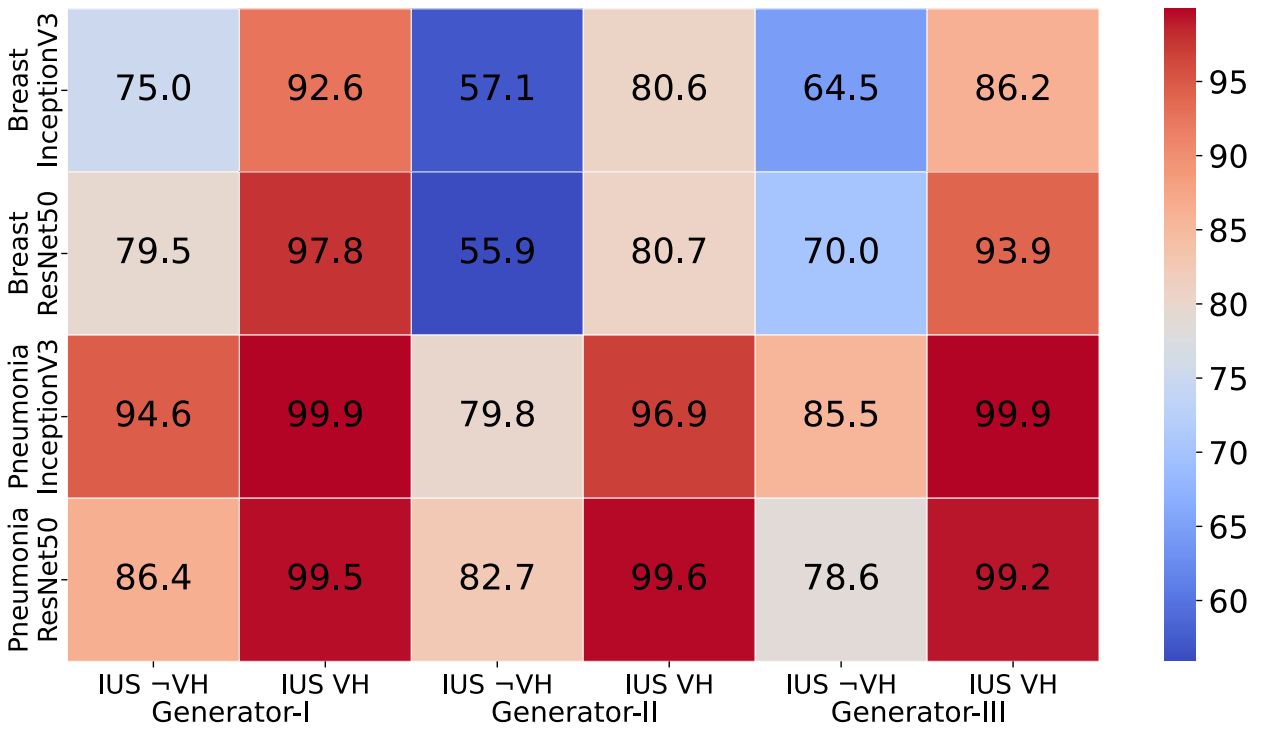

Fig. 12. Classification performance obtained after training on real grayscale datasets from the MedMNIST collection, and testing on synthetic grayscale image subsets with IUS VH and IUS ¬VH.

VH were assessed by following the exact methodology followed in Section IV.C.

In the first evaluation phase, classifiers were trained on synthetic images with IUS VH and tested on real images and their performance was compared against classifiers trained with randomly selected synthetic images, regardless of their IUS (Section IV.C.1). Table VI summarizes the classification results obtained. In the case of BreastMNIST dataset, using training images with IUS VH from *Generator-II*, InceptionV3 yielded a relative AUC improvement of 9.1% (from 82.0±5.0% to 89.5±5.2%). Resnet50 yielded a relative AUC improvement of 9.5% (from 74.9±2.6% to 82.0±2.5%) with IUS VH images from *Generator-III*. Similar improvement trends were observed even when training subset sizes were reduced to 80% and 60%, with InceptionV3 trained with 60% of the IUS VH subset from *Generator-II* scoring improvement gains of up to 15.1%. In the case of PneumoniaMNIST dataset, the IUS VH images from *Generator-II*, achieved a relative AUC improvement of 1.5% (from 88.3±1.7% to 87.0±2.3%). Experiments with the 80% of IUS VH subset from *Generator-II* resulted in relative AUC improvements of 0.7% with ResNet50.

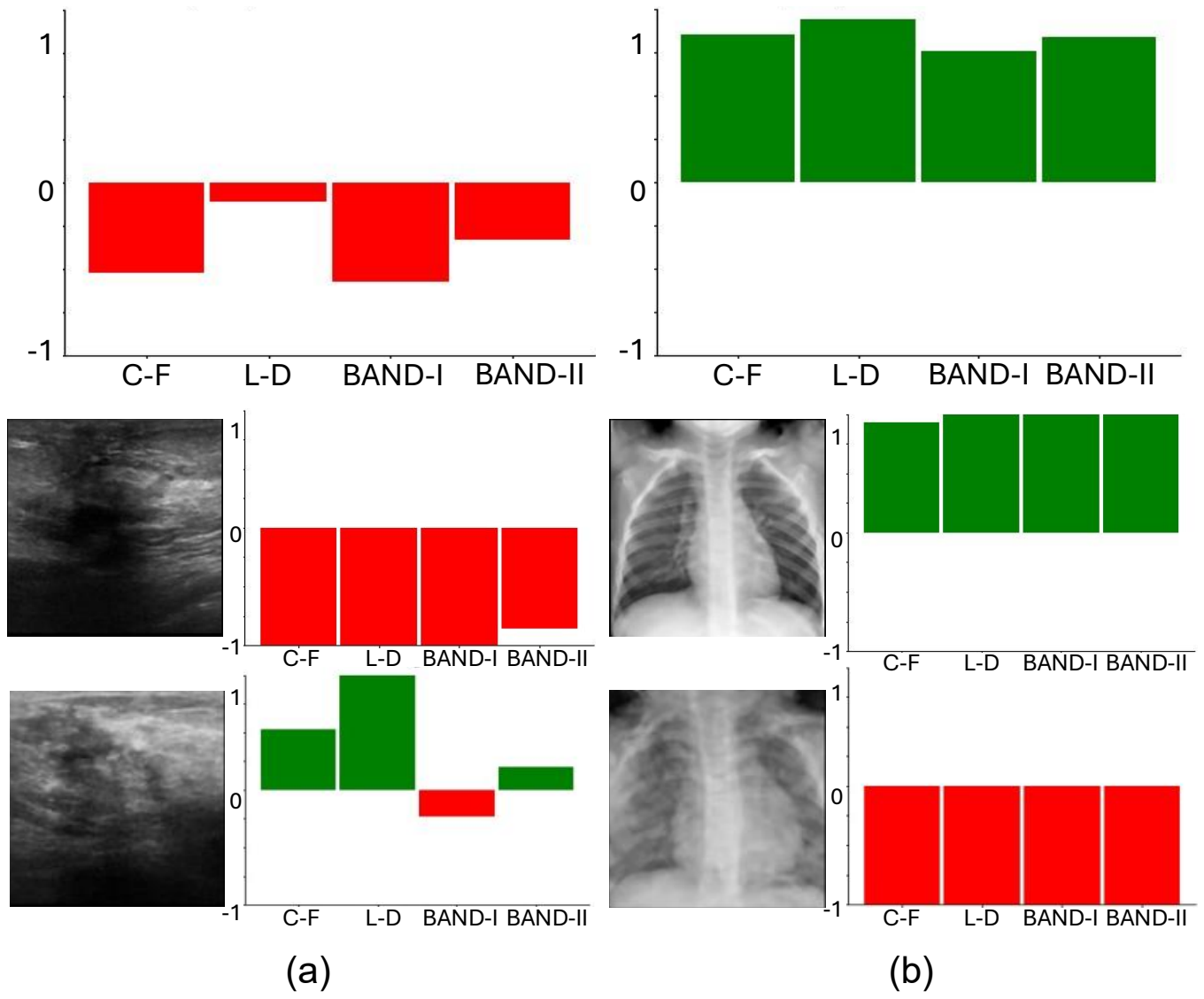

Fig. 13. Representative examples of baseline feature contribution profiles $C_b$ (row 1) for abnormal (a; red color) and normal (b; green color) classes of IUS VH (row 2) and IUS ¬VH (row 3) from the BreastMNIST and PneumoniaMNIST datasets. C-F, L-D, BAND-I and BAND-II denote the Coarse-Fine, Light-Dark, first- and second-BAND PFMs, respectively [20].

In the second evaluation phase, classifiers trained on real data were used to classify IUS VH and ¬VH synthetic subsets (Section IV.C.2). The classification results in terms of AUC scores are summarized in Fig. 12. It is shown that the synthetic subsets with IUS VH consistently achieved higher AUC scores than the subsets with IUS ¬VH, for both datasets and across all classifiers and synthesis methods. Notably, the synthetic images with IUS VH, generated from *Generator-II* and *III* based on the BreastMNIST dataset, yielded AUC improvements higher than 20.0%. The highest AUC improvement achieved was 24.8% (VH:80.7% vs ¬VH:55.9%) in the case of *Generator-II*, using ResNet50. The highest AUC improvement in the case of synthetic images generated based on the PneumoniaMNIST dataset, was 20.6% (VH:99.2% vs ¬VH:78.6%) using *Generator-III* and ResNet50.

Figure 13 showcases representative interpretation examples obtained for abnormal (Fig. 13(a)) and normal (Fig. 13(b)) images with IUS VH (in row 2), along with counterexamples, *i.e.*, images with IUS ¬VH (in row 3). It can be observed that synthetic images with IUS VH follow the trends (ascending or descending) of the baseline feature contribution profile, *i.e.*, $C_b$ (Fig. 13, row 1), whereas images with IUS ¬VH deviate strongly from $C_b$. The ultrasound image with IUS ¬VH (Fig. 13(a), row 3) demonstrates irregular lesion appearance and discontinuities in texture (C-F, PFM); the overall echogenicity reflects non-uniform tissue density (L-D PFM) with lighter areas (hyperechoic) appearing granular (BAND-II PFM). The X-ray image with IUS ¬VH (Fig. 13(b), row 3) exhibits a complete inversion of the $C_b$ PFMs that can be attributed to irregular thoracic anatomy including bone structures with discontinuities, blurred airways (vessels), granular texture and uneven lung intensity, Overall, the results obtained (Table VI, Fig.12, Fig. 13) demonstrate the generalizability of IUS measure across different imaging domains, extending its applicability beyond color medical modalities.

### H. Expanding IUS evaluation in video classification

Considering that clinical interpretation in endoscopy is usually based on a video, and not still images, we also investigated the impact of using IUS to select the most useful synthetic images for the classification of real video sequences. To this end, a publicly available endoscopic video dataset, namely PolypGen [36], was used. PolypGen comprises colonoscopic image sequences as well as still polyp images acquired across multiple clinical centers. The image sequences consist of temporally ordered frames and provide sequence-level classification labels suitable for this type of evaluation. The sequences belong to two classes, namely normal mucosa (labeled as negative) and mucosa with polyp findings (labeled as positive). The positive sequences contain polyps with various clinical characteristics, including size and type, captured from various viewpoints. In total, this subset includes 8,037 images (3,762 positive and 4,275 negative). The number of frames per sequence ranges from 15 to 250 for the positive sequences, and from 40 to 315 for the negative sequences. In the PolypGen dataset a single label is provided per sequence, while positive sequences may include frames without visible polyps.

In this case study, image classifiers were trained using synthetic images generated based on images of the PolypGen dataset, and tested on the real image sequences from that dataset. Synthetic images were generated using *Generator-I* and *-II,* which were trained separately on real still images from PolypGen. For each generator, two synthetic training image subsets were created according to Section *IV.C.1*: a subset comprising solely synthetic images with IUS VH, and a control subset comprising randomly selected synthetic images. To estimate the IUS scores for the synthetic image samples, an EPU-CNN model trained on the real PolypGen images was employed. Each synthetic image subset (IUS VH or control) was used separately to train the classifier (InceptionV3 and Resnet50). The trained classifers were evaluated using the same test set of real image sequences.

TABLE VII
REAL IMAGE SEQUENCE CLASSIFICATION RESULTS USING SYNTHETIC TRAINING DATA FROM POLYPGEN DATASET

| Data Generator | Utility | ***PolypGEN*** | | | |
|---|---|---|---|---|---|
| | | ***InceptionV3*** | | ***ResNet50*** | |
| | | Accuracy | AUC | Accuracy | AUC |
| *Generator-I* | - | **75.0 ± 10.8** | 87.3 ± 17.3 | 56.0 ± 13.6 | 62.6 ± 17.4 |
| *Generator-I* | *VH* | 74.0 ± 15.4 | **88.6 ± 14.1** | **70.0 ± 10.5** | **77.2 ± 9.9** |
| *Generator-II* | - | 61.0 ± 6.4 | 78.2 ± 12.1 | 60.5 ± 11.0 | 71.6 ± 9.7 |
| *Generator-II* | *VH* | **61.5 ± 16.3** | **80.3 ± 17.4** | **74.5 ± 9.3** | **85.3 ± 8.5** |

During the classification performance evaluation, each image within a sequence was processed independently by a trained classifier, to obtain a predicted label (positive or negative). The image-level predictions for each sequence were aggregated through a voting process to classify the sequence as positive or negative [43]. Following this approach, a sequence was classified as positive if at least $k = 5$ images in that sequence were predicted as positive. This approach yields predictions that are consistent with the sequence-level ground truth annotations provided by PolypGen dataset. While this voting process is employed to assign labels to each sequence, it lacks providing continuous scores for the classification performance evaluation in terms of AUC. Thus, following the approach described in [44], the maximum probability among the image-level predictions of a sequence, was used as the representative classification score for that sequence. This ensures that image sequences containing frames with visible polyps are assigned higher scores, while these containing solely polyp-negative frames are assigned lower scores.

Table VII summarizes the classification results obtained following a five-fold cross validation procedure. It can be observed that InceptionV3 yielded an AUC improvement of 1.4% using the IUS VH images from *Generator-I* and an AUC improvement of 2.0% using the IUS VH images from *Generator-II*. Classification with ResNet50 achieved higher improvement gains in terms of AUC. Specifically, the AUC improvement using IUS VH images from *Generator-I* was 14.6%, while the AUC improvement using IUS VH images from *Generator-II* was 13.7%.

Overall, the results of this case study demonstrate that synthetic images with IUS VH can yield gains also for classification of real video sequences. It is worth highlighting that PolypGen contains images with various clinical characteristics, *e.g.*, polyps of different sizes and types, collected across multiple clinical centers. This data variability could explain the fact that the performance of the classifiers trained with synthetic images having IUS VH did not exceed 88.6% (InceptionV3) and 85.3% (ResNet50) in terms of AUC.

## V. CONCLUSIONS

This paper presented IUS, a novel measure that answers the still open question “how can we quantitatively assess the similarity of a synthetically generated set of images or singletons to a set of real images in a given application domain.” Unlike current approaches, IUS answers this question by considering the utility of the synthetic datasets. The utility similarity was considered as a quality of practical value, which unlike fidelity and diversity, it can be used as a standalone measure to assess synthetic images. The contributions of this study can be summarized as follows: (a) IUS evaluates synthetic images based on clinically relevant perceptual features that are crucial for medical image interpretation, resulting in more reliable assessments; inspired by human perception and cognitive processes [19] , the evaluation criteria of IUS consider color and texture that are also used by medical experts for recognizing synthetic medical images [10]. (b) Unlike black-box Inception-based measures, *e.g.*, IS, IUS enables an interpretable assessment process. (c) Contrary to task-specific performance measures that can only quantify the utility of synthetic image sets, IUS can evaluate singletons, *i.e.*, individual synthetic images. (d) By accepting IUS as a standard measure to infer utility similarity of synthetic medical images, the need for time-consuming exploratory experimental procedures is mitigated. This is because IUS directly quantifies the utility similarity of images, as opposed to the conventional evaluation procedure, which requires repeatedly performing multiple classification experiments with various models and training settings to infer the usefulness of synthetic images in DL applications.

A comprehensive experimental evaluation on several publicly available benchmark datasets validated the effectiveness of IUS. The main conclusions can be summarized as follows: (a) The selection process for constructing synthetic datasets comprising synthetic images with IUS VH for training DL models leads to improved image classification performance, even with datasets of smaller size; indicating a consistent generalization performance. (b) Synthetic medical images with IUS VH exhibit equivalent quality with that of real images. (c) Synthetic images with IUS VH provide better FID scores suggesting correlation between utility similarity and visual fidelity, narrowing the existing gap in synthetic medical image evaluation. (d) Synthetic data diversity remains approximately unaffected over different IUS levels. (e) IUS can be applied to assess images obtained by any synthetic image generation model, such as GAN, VAE, or LDM, using various medical imaging modalities, ensuring broad applicability in clinical settings. (f) Although IUS is designed for evaluating still synthetic images, the results from a case study using endoscopic image sequences indicate that synthetic images with IUS VH can support the training of classifiers for lesion detection in video endoscopy. A limitation of that study is that IUS is used per frame, disregarding the temporal relationships between consecutive video frames.

The study focused on color medical images where the interpretability of IUS can fully unfold its potential. However, as demonstrated in Section IV.G, it can be easily adapted for grayscale images, such as radiology images, with the use of different PFMs. Future research perspectives include investigation on complementary PFMs that encode image characteristics, such as shape and/or morphology of the depicted anatomical structures and extending the generalization of IUS by incorporating knowledge from various medical imaging modalities and datasets, including datasets captured from different imaging systems, into a single IUS instance. Moreover, future research could involve extension of IUS in the context of medical video generation, by incorporating a temporal dimension to similarity assessment, and extending its evaluation on other video datasets, *e.g.*, [45], [46]. Also, it should be highlighted that IUS is generic; therefore, its application for the evaluation of non-medical synthetic images in other domains, such as industrial inspection, is also a promising perspective.